\documentclass[journal,twoside]{IEEEtran}
\usepackage{amsmath,amsfonts}
\usepackage{url}
\usepackage{verbatim}
\usepackage{graphicx}
\usepackage{cite}
\usepackage{caption}
\usepackage{subcaption}
\usepackage{soul}
\usepackage{diagbox}

\begin{document}
%
\title{Stochastic Modeling of Road Hazards on Intersections and their Effect on Safety of Autonomous Vehicles}
%
%
%

\author{Peter Popov$^{1}$,
        Lorenzo Strigini$^{1}$,
        Cornelius Buerkle$^{2}$,
        Fabian Oboril$^{2}$,
        Michael Paulitsch$^{2}$,
\thanks{$^{1}$City St George's, University of London, $^{2}$ Intel Deutschland GmbH}
}

\markboth{IEEE TRANSACTIONS ON INTELLIGENT TRANSPORTATION SYSTEMS}%
{P. Popov et al.: Stochastic Modelling of Road Hazards}

%



\maketitle

\begin{abstract}
Autonomous vehicles (AV) look set to become common on our roads within the next few years. However, to achieve the final breakthrough, not only functional progress is required, but also satisfactory safety assurance must be provided. Among those, a question demanding special attention is the need to assess and quantify the overall safety of an AV. Such an assessment must consider on the one hand the imperfections of the AV functionality and on the other hand its interaction with the environment. 
In a previous paper we presented a model-based approach to AV safety assessment in which we use a  probabilistic model to describe road hazards together with the impact on AV safety of imperfect behavior of AV functions, such as safety monitors and perception systems. With this model, we are able to quantify the likelihood of the occurrence of a fatal accident, for a single operating condition. 
In this paper, we extend the approach and show how the model can deal explicitly with a set of different operating conditions defined in a given ODD.

\end{abstract}


%
\IEEEpeerreviewmaketitle

\section{Introduction}
%
%
%
%
\IEEEPARstart{A}{ssessing} 
 safety of autonomous vehicles (AV) poses new challenges. The long-standing approaches to vehicle safety assessment, related to functional safety~\cite{ISO26262}, are known to be problematic when dealing with functions based on machine learning. Consequently, new alternatives are evolving, among them is the standard on “Safety of the Intended Functionality (SOTIF)” (ISO 21448)~\cite{ISO21448}, the new safety standard for AV decision making~\cite{IEEE2846}, or the newly-released standard of safety of AI components in vehicles~\cite{ISO8800}. In addition, national authorities are currently setting up the required legislative boundaries for the public use of AVs~\cite{DiFabio2017}. While these standards define "guardrails" within which manufacturers must operate, an important question remains: How can an AV be proven to be safe enough for certification to participate in general traffic on public roads?  

To demonstrate that an AV is safe enough, manufacturers need to demonstrate, as a minimum, that the rate of accidents will be below a required bound. Targeting the average safety performance of a human driver (in Germany, as an example) would mean that the accident rate must be below $10^{-5}$ per hour~\cite{DiFabio2017}. Realistically, the target will be to exceed the safety performance of good drivers under various operating conditions, so the above numbers indicate a bound on acceptable orders of magnitude. The obvious solution of deriving safety arguments from the evidence of large amount of safe driving by AVs~\cite{Kalra2016} is impractical, in view of the limited amount of trial driving that is considered affordable (in terms of cost and, possibly, of risk to the public) before the intended mass deployment~\cite{Kalra2016,Littlewood1993}. To reach high enough confidence in a vehicle to allow its mass deployment in commercial/consumer use, one needs a basis of confidence, prior to observing extensive safe driving. This prior confidence would require a sound approach that can be either mathematically verified or anyway does not require such extensive new data for validation.  

To demonstrate how prediction of accident rates supports deriving required reliability of specific functions in an AV, we introduced~\cite{Buerkle2022} an approach to modeling how the probabilities of various malfunctions in an AV contribute to the frequency of accidents. We addressed the key questions raised above by a formal approach to safety assessment. We proposed a comprehensive model to estimate the probability of accident ("catastrophic failure” in the terminology of our models) of an AV over a period (e.g., 1000 h) of driving, which implemented the above approach of considering road hazards together with failures in the AV processing pipeline (e.g., perception or planning). For this purpose, we modeled the occurrences of failures as non-homogeneous stochastic processes and demonstrated that many of the model parameters could be estimated using empirical data from actual driving. Among these parameters are the parameters of road hazards – such as the intervals between their occurrence, and their durations, as recommended by ISO 26262, and the duration of perception system failures, e.g., to detect an obstacle on the road. 
In summary, our proposed model captures SOTIF performance limitations concerning situational awareness and their impact at system level, which can be used early in the design process to help in making appropriate design decisions. At the same time it contained some simplifications. 

In this paper we extend the previously developed model by finer detailing, to account for: (a) different driving conditions – driving through road intersections and over roads between intersections; (b) different speeds of AV driving – fast and slow. 
(c) different severity levels of accidents. These extensions are methodologically important as they allow us to accommodate the concept of Operational Design Domain (ODD)~\cite{BSI2020}, which has been recognized as a significant factor affecting AV safety and therefore has to be accounted for in safety assessment. 
Indeed our example of use of our enhanced model exemplifies how safety can vary between the operating conditions of an ODD.
Furthermore, the evaluation also reveals how the contribution to the overall safety can vary significantly among the different components of an AV.

The paper is organized as follows: Section II defines the essential concepts used in the proposed model. The model itself is presented in Section III. Section IV covers our findings derived by solving the model, followed by a discussion in Section V and related research in Section VI. Finally, the paper is concluded in Section VII.

\section{Concept}

Our purpose is to model how failures of an AV's functions interact with road hazards, and sometimes cause accidents. Thus, in the model an accident can only be triggered by a hazard on the road. A hazard is a temporary state of the environment~\cite{ISO26262}. Once it occurs, it will have a duration (e.g., a few seconds, possibly longer) after which it will cease to exist. 

\begin{figure}[t!]
    \centering
    \includegraphics[width=0.6\linewidth]{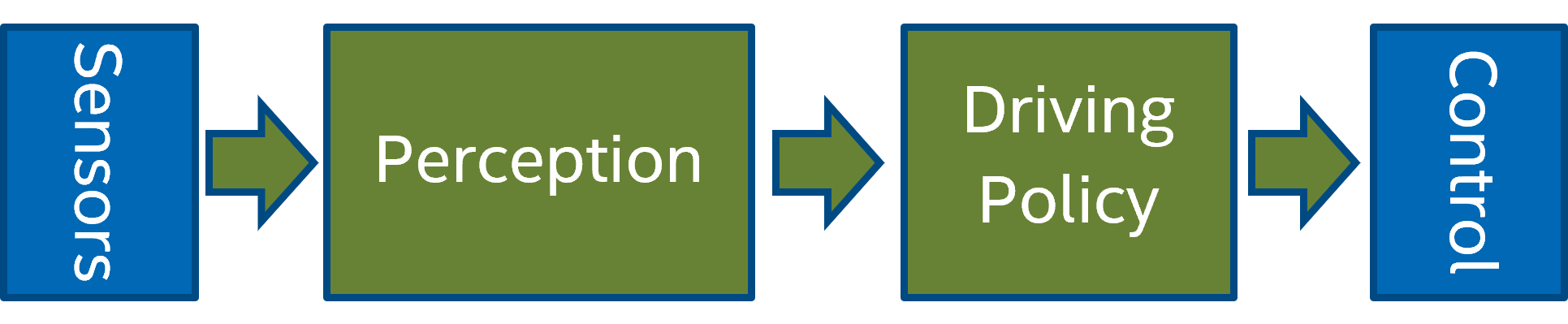}
    \caption{AV system architecture. Both perception and driving policy can fail when responding to road hazards, and thus cause accidents.}
    \label{fig:AVarch}
    \vspace{-0.4cm}
\end{figure}

 A road hazard is a phenomenon external to the AV: its occurrence is not affected by whether the hazard is recognized correctly by observers, e.g., the AV's perception system, or not.  Examples of road hazards would be unusual situations on the road (e.g., due to weather conditions, intense traffic, vehicles ahead braking, etc.). These occur randomly, and create only the potential for accidents, but the escalation of hazards to accidents is non-deterministically characterized by the likelihood of the vehicle failing to correctly respond to the hazardous situation. For an AV, there are two main sources for such a failure, as illustrated in the system architecture in Figure~\ref{fig:AVarch}. First, the perception system could fail to detect the hazard. Second, the driving policy could fail to respond properly even with perfect perception.  Both of these cases are covered by our proposed model and are discussed in the evaluation section. In this regard it is worth noting that there are various safety concepts for both aspects (e.g.,~\cite{shalev2017formal,buerkle2022monitor,oboril2021rss+}), yet no concept guarantees freedom from accidents.
 These often involve dedicated safety mechanisms (called e.g. ``safety monitors'', or ``safety guards'') to specifically detect and interrupt accident sequences despite possible failure of other functions. 
 Without loss of generality, we refer in this work only to the high-level components: perception and driving policy.

\begin{figure}[b!]
    \centering
    \vspace{-0.4cm}
    \includegraphics[width=1.\linewidth]{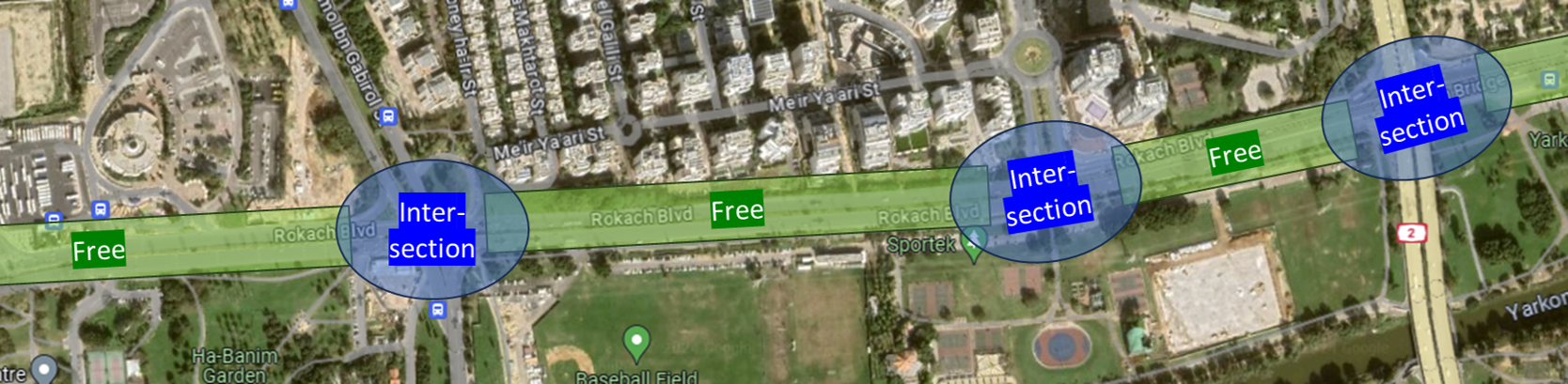}
    \caption{Example driving scenario: Multilane road with several intersections}
    \label{fig:road}
\end{figure}

In this work, we use a formal model to describe the stochastic processes of randomly occurring hazards and the ability of the vehicle to correctly respond. To capture a variety of different traffic situations common for urban or highway driving, we use in this work a multilane driving scenario that also includes various intersections. An example of such a road can be seen in Figure~\ref{fig:road}.

\begin{figure}[t!]
    \centering
    \includegraphics[width=0.5\linewidth]{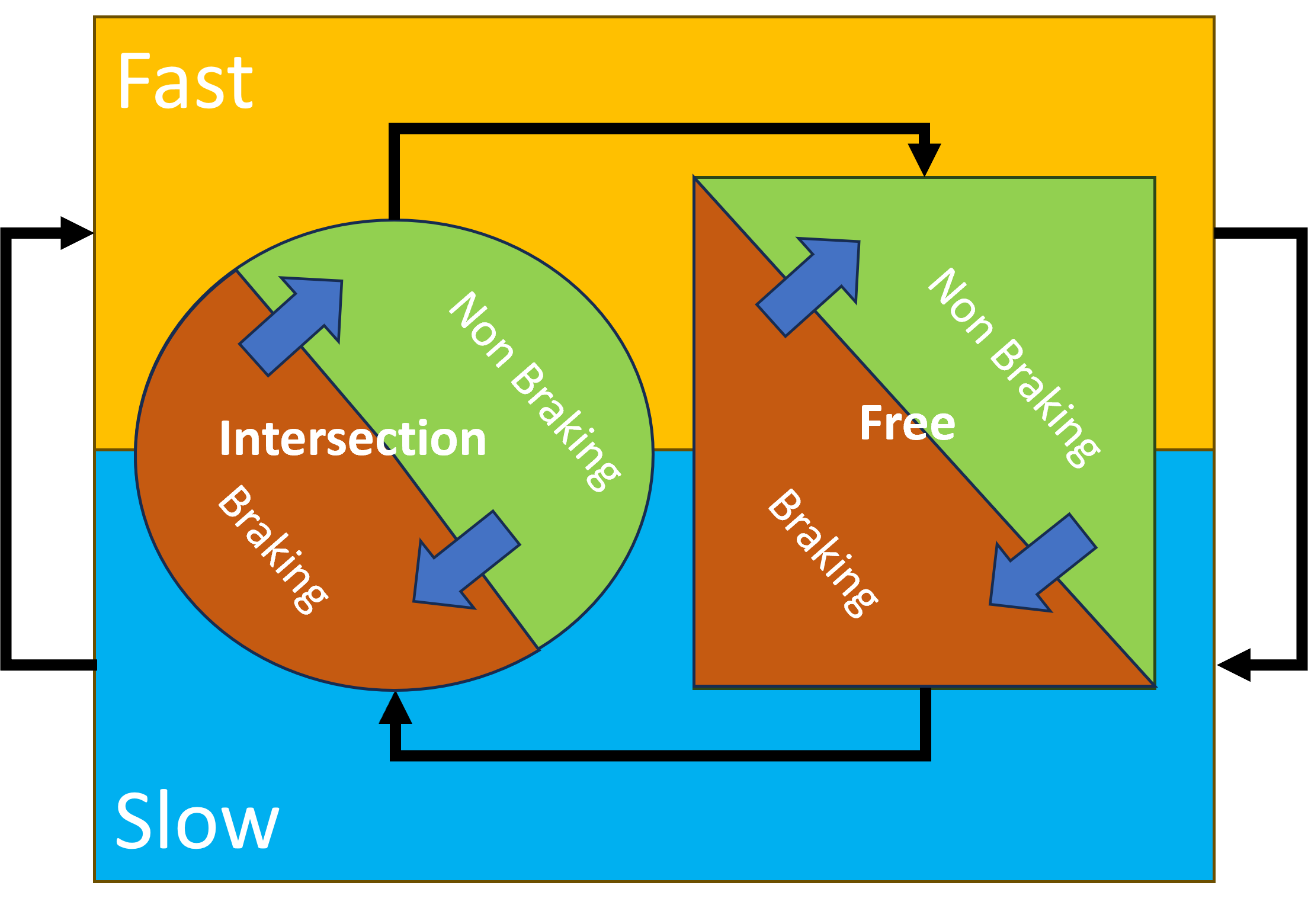}
    \caption{States of the ego vehicle.   Vehicle transitions between \textit{free} driving and \textit{intersection}. In parallel it is responding to the environment, which at times requires braking to respond to hazards. The driving speed is captured by the states \textit{fast} and \textit{slow}.}
    \label{fig:states}
    \vspace{-0.4cm}
\end{figure}

\begin{figure}[b!]
    \centering
    \vspace{-0.4cm}
    \includegraphics[width=1.\linewidth]{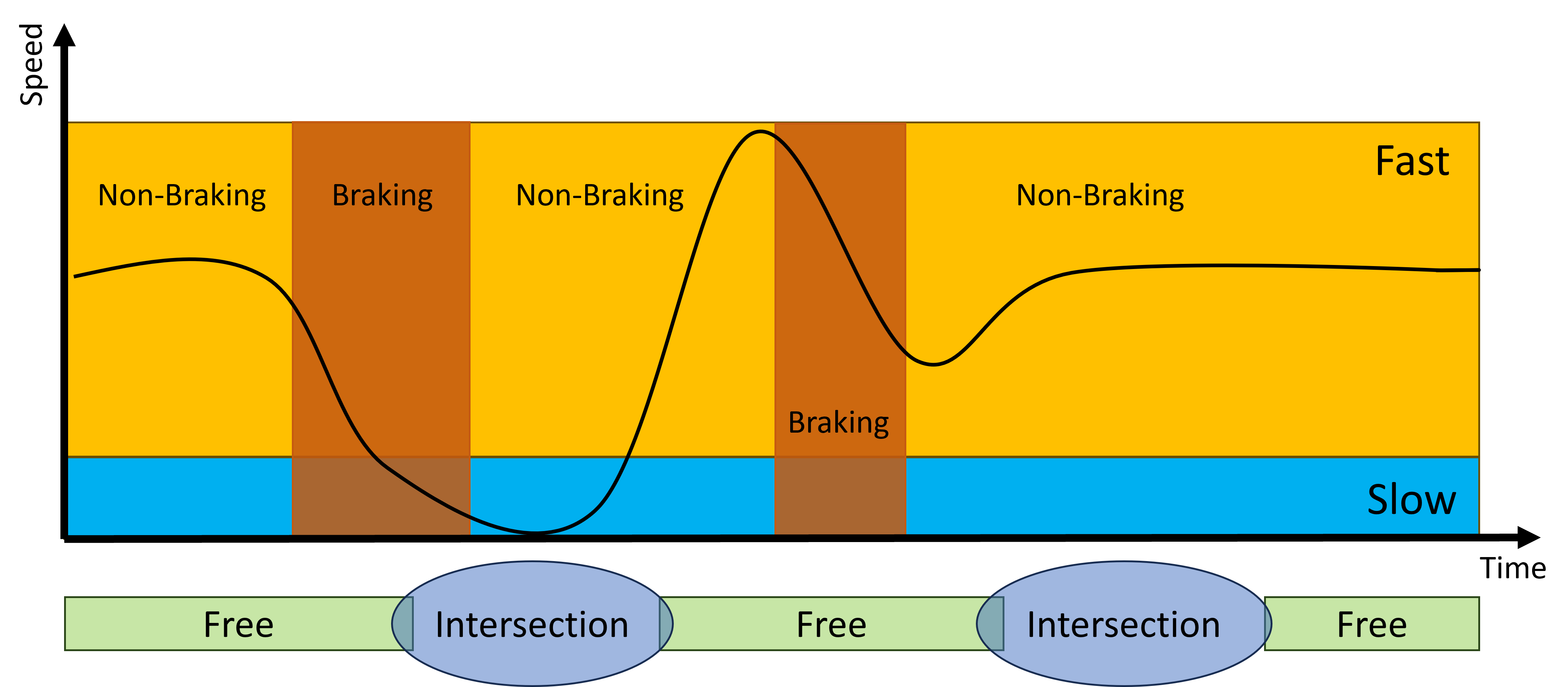}
    \caption{Example speed diagram of a car driving in the modeled scenario, with an overlay of states from Fig. \ref{fig:states}. The drive consists out of sections that require braking to hazards and which that don't. It can be seen that deceleration can happen as well in a non-braking situation. - This is a hypothetical segment from the kind of data we have used, illustrating how the model probabilities and transition rates can be derived from the data.  }
    \label{fig:Velocity}
\end{figure}

In this use case, a generic driving mission can be described in terms of two situations:
the car is either driving within an intersection region or driving “freely” on a link between intersections. Consequently, the road segment, the car is currently driving on, can be described by two states \textit{free} and  \textit{intersection} where ``intersection'' denotes driving through a road intersection, and ``free'' denotes driving 
on roads between intersections. The car will transition between these two states, and our simulation will assess how long this will continue, across a sequence of trips,  until stopped by an accident or completing without an accident the driving mission (e.g. of a certain number of hours). 
The driving speed of the car can be regarded as a second state variable. Here, we distinguish between two possible states \textit{fast} and \textit{slow}, where ``fast'' indicates that the speed is high enough to cause serious accidents, while ``slow'' indicates driving at low speed where accidents will not cause serious injuries. Assuming only accidents between 4-wheeled vehicles, for the threshold between slow and fast we use 15m/s: according to statistics (e.g., ~\cite{Oboril2022,2019NHTSAAccidents,JUREWICZ20164247}), above this speed there is a strong likelihood for severe injuries for passengers in car-to-car collisions. Figure~\ref{fig:states} and Figure~\ref{fig:Velocity} show the relationships between the states and the possible transitions. 

Furthermore, the car needs to respond to the surrounding environment by adapting its speed. From a safety point of view, we distinguish two situations: \textit{braking} and \textit{non-braking}. ``Braking'' indicates a potentially presence of an hazard,  where the ego vehicle needs to brake, e.g., it must adjust its speed to a leading car driving at lower speed. 
A failure in accomplishing this response might result in an accident, unless the hazard is not resolved on its own, e.g. by acceleration of the leading car. Such a failure can be caused by an error in the vehicle perception system or driving policy component.

In summary, we have identified, so far, three attributes of the state of a car and its environment: road condition (``Free/Intersection''), speed  (``Fast/Slow''), presence of hazard (``Braking/Non-Braking''), and thus 8 possible states.

We use a SAN (Stochastic Activity Networks) model~\cite{WH2001} to capture all these possible states and to analyze the impact of potential errors in the vehicles AV system. All the models in this paper were specified and solved using the Mobius tool\footnote{https://www.mobius.illinois.edu/}.

\section{Model}

The model itself is built hierarchically, i.e., its behavior in each one of the 8 high-level states explained in the previous section is further detailed in an ``atomic model'' (in SAN terminology) for that state. The complete SAN model includes many possible transitions among the possible states. Explaining each detail would go beyond the limits of this paper, therefore, we skip them here, but interested readers can find these details and have full access to the model~\cite{SANModel2024}.

\subsection{High-Level Model}

\begin{figure}[t!]
    \centering
    \includegraphics[width=0.8\linewidth]{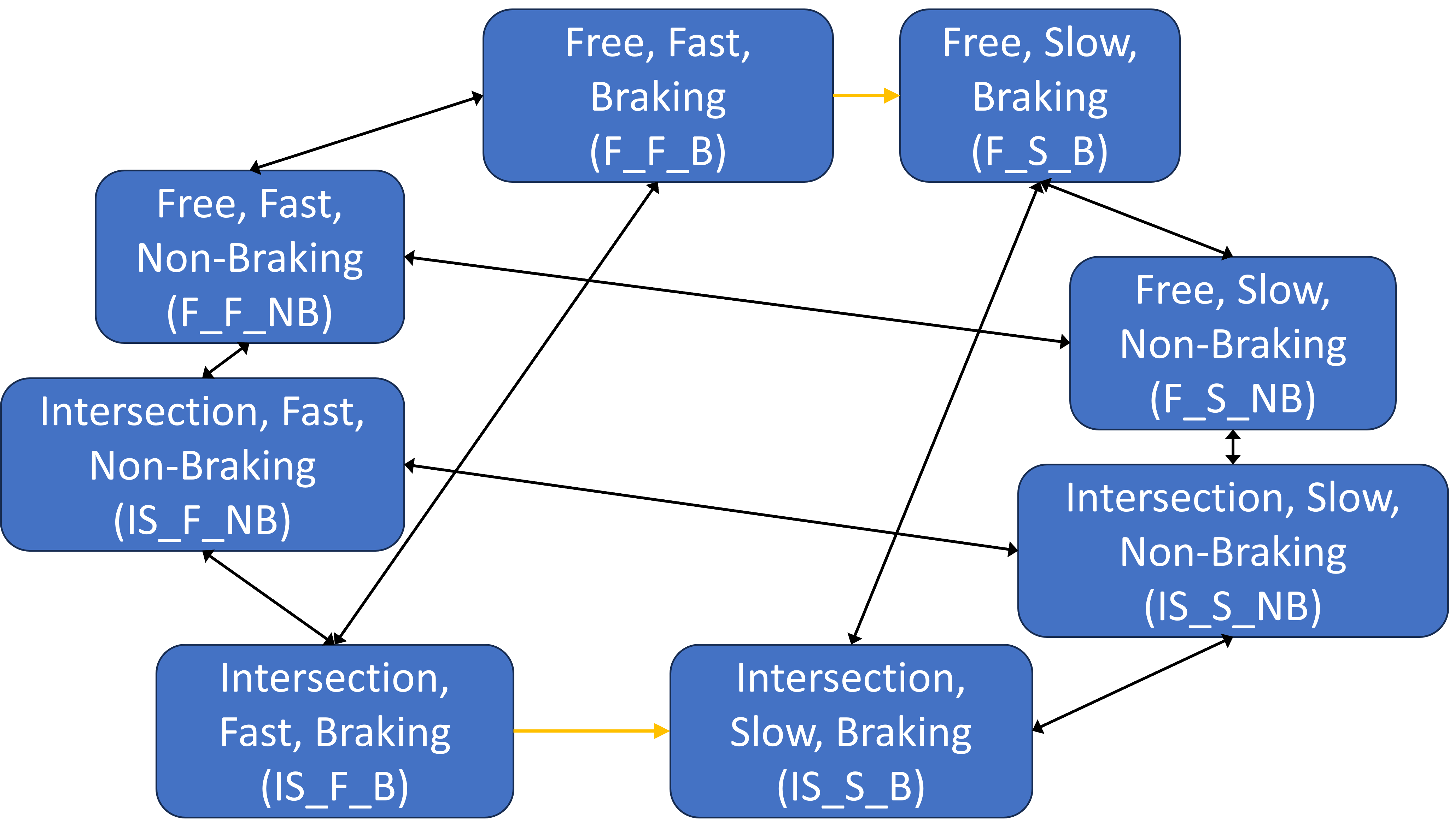}
    \caption{The high-level states of the model and possible transitions between them. For simplicity this figure only shows 1st-order transitions (i.e. only one attribute changes value) omitting 2nd-order and 3rd-order ones.}
    \label{fig:HighModel}
    \vspace{-0.4cm}    
\end{figure}

The top-level states describe the road condition, speed and situation of the vehicle. A road trip can then be seen as a sequence of visits to these states (as shown in Figure~\ref{fig:HighModel}), determined by a random process. The vehicle will spend in each visited state a certain amount of time (\emph{sojourn \footnote{sojourn = a short stay somewhere}
time}), which is a random variable, and then move to one of the other 7 states, with probabilities that are described as model parameters. Thus, the movement of the vehicle is modeled – as far as the possibility of accidents is concerned – as a random process of visiting the states illustrated in Figure~\ref{fig:HighModel} and spending in each of them a random period of time. The transitions will have different probabilities, reflecting the behavior of vehicles in real traffic. However, some transitions are physically impossible, e.g., if a car drives slowly and is currently braking, it is impossible to change to fast driving. Consequently, some entries in Table~\ref{table:state_transition} (in the Appendix) show 0 frequency in our data set.


\begin{figure}[b!]
    \centering
    \vspace{-0.4cm}
    \includegraphics[width=0.8\linewidth]{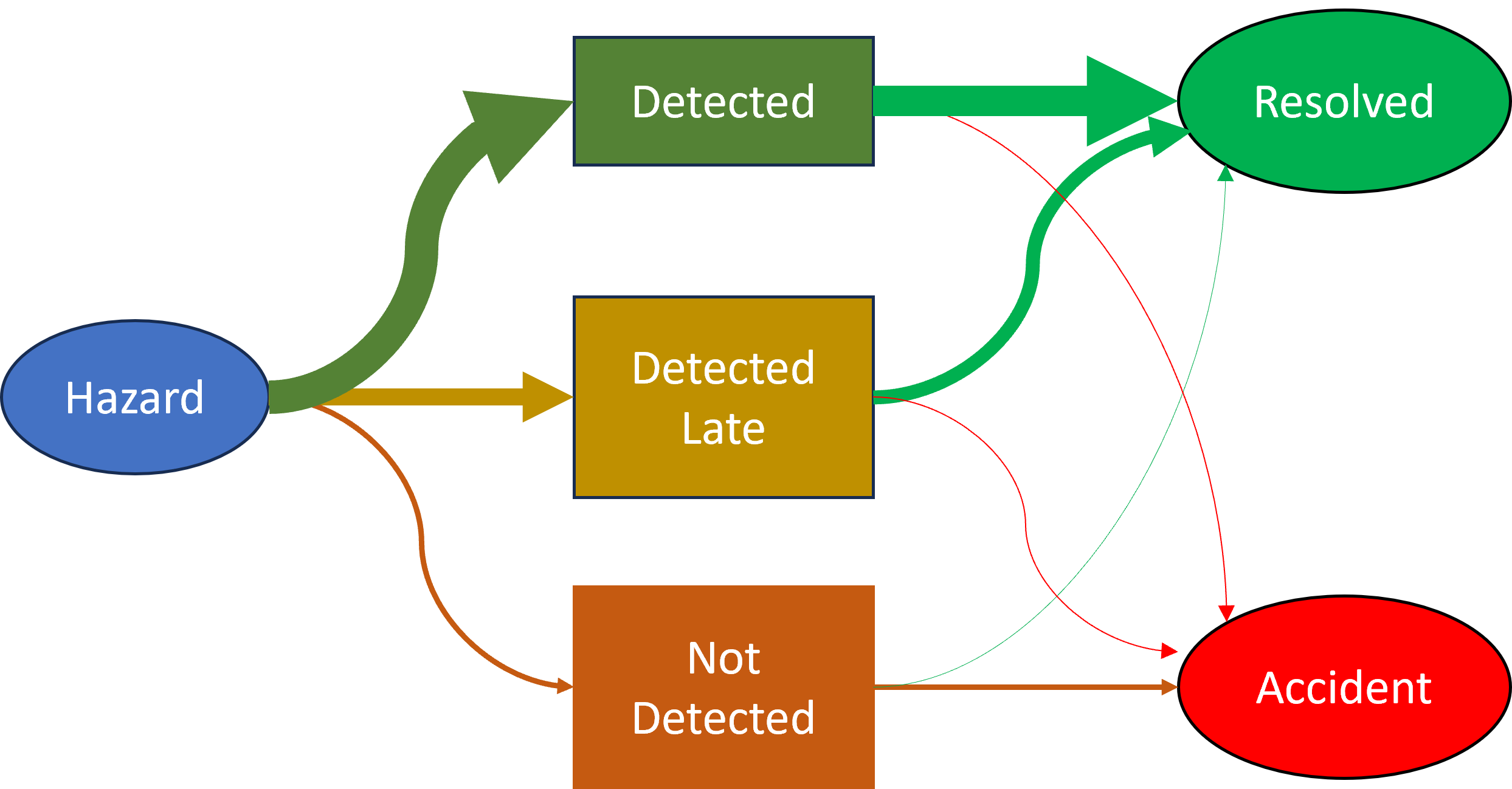}
    \caption{Possible transitions in a hazardous state. The different arrow widths reflect the different path probabilities (more in Section~IV)}
    \label{fig:HazardAccident}
\end{figure}

\subsection{Atomic models}
The evolution of hazards, and the vehicles reaction to them, are modeled through \emph{atomic sub-models} of the high-level states. These atomic [sub]models consist of their own sub-states with transitions between them.
\subsubsection{Hazardous states}
 In this regard, the \textit{braking} high-level states are of particular interest.
They represent presence of a road hazard requiring the vehicle to decelerate, e.g., a road occupant preceding the ego vehicle braking. The ego vehicle must detect this road hazard and react correctly. 
We consider the following scenarios:
\begin{itemize}
    \item the road hazard has been \textit{detected} correctly and on time;
    \item the road hazard has been overlooked (i.e., \textit{not detected});
    \item the road hazard has eventually been detected after a short period of temporarily overlooking it (i.e., \textit{detected late}).
\end{itemize}

The transitions among the sub-states of the atomic models for these conditions are illustrated in Figure~\ref{fig:HazardAccident}. 
If not properly detected, a hazard has a high likelihood to lead to an accident. However, some hazardous situations will resolve themselves, even if not detected properly; but there is also the chance that some hazards will end up in an accident despite being detected, if the vehicle cannot resolve the situation on its own (e.g., a close cut-in by another vehicle). All of these potential sequences are captured by our model.

Each of the high-level hazardous states (that is, the 4 high-level states with braking), as detailed in its corresponding atomic submodel, contains a special sub-state \textit{Accident}. 
This is an ``absorbing'' state: once it is entered, no more transitions occur and a simulation stops here. 
We distinguish between:
\begin{itemize}
\item a \textit{serious Accident}, which can only occur when the vehicle is traveling fast,
i.e., in hazardous states F\_F\_B and IS\_F\_B.
\item 	a \textit{minor Accident}, which can only occur when the vehicle is traveling slowly: 
hazardous states F\_S\_B and IS\_S\_B.
\end{itemize}

\subsubsection{Non-hazardous states}
We model a single type of hazard state, called ``braking'' since the required action is braking.
Later in the paper we will discuss the implications of relaxing this model assumption. Until then it holds that all and only the  \textit{non-braking} states are non-hazardous states. The only transitions from them are to other high-level states, never to Accident states.

\begin{figure}[t!]    
    \centering    
    \includegraphics[width=0.95\linewidth]{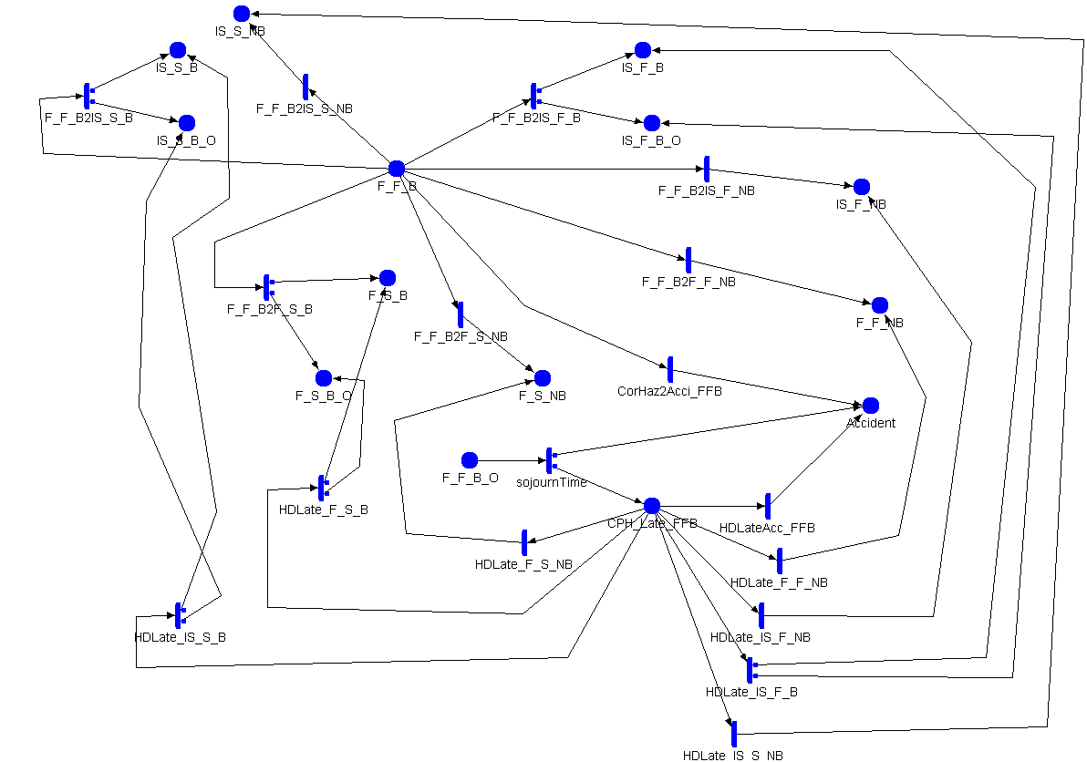}
    \caption{Atomic SAN model for the high-level state F\_F\_B (fast, free, braking)}
    \label{fig:atomicSANFFB}
    \vspace{-0.4cm}
\end{figure}

\subsubsection{Transitions}
Each transition from a high-level state is modeled 
as a ``timed activity''. Each timed activity is defined with a probability distribution for the random time it takes for the transition to ``fire'', i.e., the time from entry into the state  until this transition to a new state takes place.
If the transition arc in the model graph branches (there is more than one possible destination state when that transition fires), ``case'' probabilities also need to be specified for each branch.

So, the vehicle will spend a random time (drawn from the distribution defined for the respective timed activity) in a certain state, and then 
transition to another state. If case probabilities are attached to the timed activity, then, once it "fires", the next state is picked at random according to the case probabilities.

We call a simulation run of the model a ``mission'', according to the Mobius terminology.
So, a mission is
executed as 
a journey through the states that lasts until a pre-set mission duration is reached or an accident takes place. Hence, a driving mission can have three outcomes:
success, if the duration of the mission is reached without entering an accident state;
a \textit{serious Accident}; or a \textit{minor Accident}.

\subsubsection{Bringing all together}
Combining all aspects summarized above is achieved via a set of atomic models for each high-level state, linked to work together by "sharing" some of the sub-states in the atomic models among high-level states, a mechanism 
available in Mobius. 
An example  atomic model, for the state F\_F\_B, is illustrated in Figure~\ref{fig:atomicSANFFB}. As one can see, the atomic model also contains the sub-state of an overlooked hazard (F\_F\_B\_O) and a sub-state for a late detection (CPH\_Late\_FFB). There are also various end points ("places") included in the atomic model. The most important one is the Accident place, but also places which are shared with other atomic models can appear. For instance, IS\_F\_B is a place shared with the atomic model with the same name used to model the transition from the high-level state F\_F\_B to the high level state IS\_F\_B (all details on shared places used to link the atomic models can be found in \cite{SANModel2024}).

\begin{figure}[t!]    
    \centering        
    \includegraphics[width=0.95\linewidth]{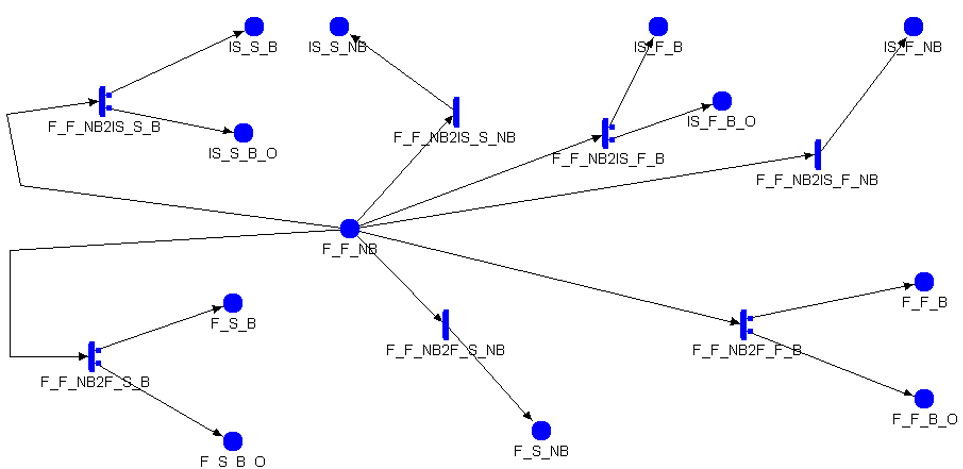}
    \caption{Atomic SAN model for the high-level state F\_F\_NB (fast, free, non-braking)}
    \label{fig:atomicSANNFFB}        
    \vspace{-0.4cm}
\end{figure}

Another example of an atomic model is depicted in Figure~\ref{fig:atomicSANNFFB}. Here, it is a non-braking state and as it can be seen in the figure, there is no accident end point present in this model, as we consider this state to be non-hazardous. This atomic model captures the sojourn time the system spends in it, and the transitions to the other high-level states using shared places. For instance, the place F\_F\_B used in the atomic model is a shared place, which also appears in the atomic model F\_F\_B shown on Figure~\ref{fig:atomicSANFFB}.   

\begin{figure}[b!]
    \vspace{-0.4cm}
    \centering
    \includegraphics[width=0.9\linewidth]{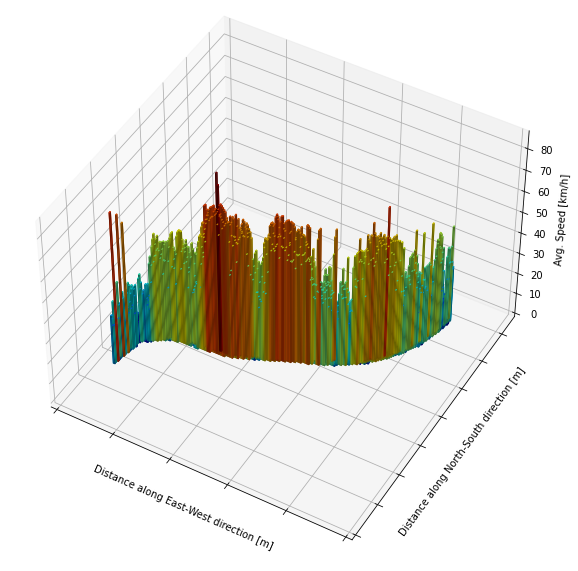}
    \vspace{-0.2cm}
    \caption{Speed distribution for a given road section}
    \label{fig:VelocityDistribution}    
\end{figure}

\subsection{Model Parameters}
\subsubsection{Transitions between  high-level states}
~\\Most of the parameters used in the model define the transitions among the 8 states. The values of these parameters were estimated using \textit{naturalistic} driving datasets about human driving behavior. These datasets include data from human-operated vehicles such as speed, accelerations, lane changes or turning behavior, leading or following vehicle ids, etc. One example dataset for German highways is HighD~\cite{krajewski2018highd}. In this work, we use a proprietary naturalistic dataset comprising driving on multi-lane roads connected through complex intersection geometries, with vehicle speeds up to 80 km/h. An example showcasing speed information from this dataset is given in Figure \ref{fig:VelocityDistribution}. While the speed category (fast or slow) can be directly retrieved from such datasets, the driving type (free or intersection) has to be obtained through location data in correlation with map information. Finally, the braking event (brake or nobrake) depends on the vehicle's deceleration in the dataset. However, as a vehicle can decelerate without braking, we assume in this work that a braking state requires at least a deceleration of 2\,m/s$^2$, in accordance to the minimum braking parameter of the IEEE 2846 standard~\cite{IEEE2846}.

While researchers often need to rely on a small public dataset, and this one suffices to illustrate our method, it is a fair assumption that the required information will be available on a large scale for automotive companies, since 
such data are already collected by many automotive companies, like Mobileye or Tesla.

Table~\ref{tab:exampleData} illustrates the content of our evaluation dataset, for a given vehicle. The rows of Table~\ref{tab:exampleData} show the duration of each \textit{state} measured in number of frames taken by an observing sensor. For instance, the first row indicates that the vehicle spent 48 frames in F\_F\_NB (free, fast, no-brake) before moving to the state F\_F\_B (free, fast, brake), etc. This data item is a realization of the transition F\_F\_NB2F\_F\_B. In this regard, it is worth recalling that in this work speeds faster than 15m/s are classifies as ``fast''. 


Following this pattern, the dataset was processed to extract all realizations within it of the 56 possible transitions  between the high-level states. A summary of these transitions is provided in Table~\ref{table:state_transition_short} and the entire list is shown in Table~\ref{table:state_transition} in the Appendix. Each row represents a transition (a timed activity in the SAN model) together with the number of times that transition occurred in the dataset. From these numbers, the rate of each transition was estimated assuming that the observed instances were sampled from a timed activity with an exponential distribution. 

\begin{table}[t!]
    \centering
    \begin{tabular}{|c|c|c|c|c|}
        \hline
        Frame \# & Driving Type & Speed & Braking Event & 
        Duration \\ \hline 
        0   & free & fast & nobrake & 48\\
        48  & free & fast & brake   & 19\\
        67  & free & slow & brake   & 27\\
        94  & free & slow & nobrake & 21\\
        115 & intersection & fast & nobrake & 18\\
        133 & free & fast & nobrake & 18\\
        \hline
    \end{tabular}
    \caption{Example data from our evaluation dataset}
    \label{tab:exampleData}    
    \vspace{-0.4cm}
\end{table}

\begin{table}[b!]
\vspace{-0.2cm}
\centering
\begin{tabular}{|c|@{\,}c@{\,}|c|c|}
\hline
Transition & Count \# & Model variable & Rate [h$^{-1}$] \\
\hline
IS\_F\_NB2IS\_S\_B & 43 & rate\_IS\_F\_NB2IS\_S\_B & 936.29 \\
IS\_F\_NB2IS\_S\_NB & 52 & rate\_IS\_F\_NB2IS\_S\_NB & 1109.88 \\
IS\_F\_NB2IS\_F\_B & 2 & rate\_IS\_F\_NB2IS\_F\_B & 1800 \\
IS\_F\_NB2F\_S\_B & 137 & rate\_IS\_F\_NB2F\_S\_B & 560.88 \\
IS\_F\_NB2F\_S\_NB & 115 & rate\_IS\_F\_NB2F\_S\_NB & 918.64 \\
IS\_F\_NB2F\_F\_B & 163 & rate\_IS\_F\_NB2F\_F\_B & 963.55 \\
IS\_F\_NB2F\_F\_NB & 2599 & rate\_IS\_F\_NB2F\_F\_NB & 948.25 \\
F\_F\_NB2F\_S\_B & 643 & rate\_F\_F\_NB2F\_S\_B & 252.79 \\
F\_F\_NB2F\_F\_B & 2870 & rate\_F\_F\_NB2F\_F\_B & 170.61 \\
\hline
\end{tabular}
\caption{Example model parameter values extracted from our evaluation dataset. }
\label{table:state_transition_short}
\end{table}

\subsubsection{Transitions within hazardous states}
~\\The probability of
overlooking a hazard and how long it is overlooked are directly linked to the quality of the perception system. In our previous paper~\cite{Buerkle2022} we explained how these values can be estimated using a publicly available dataset such as  Lyft~\cite{Transport2021}. Car companies will have a good understanding about these quality measures for their systems based on their own data collection activities.
Having this data, one can derive the parametrization for the \textit{sojournTime} present in each atomic model for braking states. For example, in case of the F\_F\_B atomic model shown in Figure~\ref{fig:atomicSANFFB}, the sojournTime reflects the transition from an overlooked hazard to either an Accident or to the sub-state CPH\_Late\_FFB. According to the data from~\cite{Buerkle2022}, the value of this parameter is 4500. The values for other atomic models are defined in Table~\ref{table:sojournTime}.

\subsubsection{Unknown transition parameters}
Other parameters, like the rate of transition to an accident (e.g. of timed activity CorHazAcci\_FFB), might be impossible to measure  accurately. Therefore, we 
use the model to perform \textit{sensitivity analysis}, i.e. conduct various experiments for a range of parameter values to assess how sensitive the model behavior is to the value of each parameter. This will produce two indications: 
(i) whether the particular parameter affects significantly the  accident probability, and (ii) if it does, establish a critical upper bound for the parameter value beyond which the probability of accidents becomes unacceptably high. 

\begin{table}[t!]
\centering
\begin{tabular}{|c|c|}
\hline
sojournTime & Rate [h$^{-1}$] \\
\hline
F\_F\_B & 4500 \\
F\_S\_B & 2250 \\
IS\_F\_B & 2250 \\
IS\_S\_B & 1225 \\
\hline
\end{tabular}
\caption{sojournTime parameter value for different atomic models derived according to~\cite{Buerkle2022}}
\label{table:sojournTime}
\vspace{-0.4cm}
\end{table}
\section{Evaluation}

In this section, we discuss the findings recorded using the model defined in the previous section. We study several different scenarios, which allow us to illustrate the insights the model can provide. A summary of all evaluations can be found in Table~\ref{tab:studyoverview}.

Each of the studies includes several ``experiments'', which define different values for the model parameters: by comparing their results we can assess the impact of model parameters on AV safety. 
The model is solved for each of the experiments (i.e. for the set of parameter values) using transient solvers, a numerical method which computes the probability of reaching the absorbing state \textit{Accident} (serious or minor) by a set ``mission time''. By varying the mission time we can trace the probability of accident over time. The transient solvers are provided by the Mobius tool. 

\begin{table}[b!]
    \centering
    \begin{tabular}{|c|c|l|}
    \hline
       Study  & Experiments & ~~~~~~~~Explanation\\ \hline
       Study 1  & Experiment 1 & Everything perfect (baseline experiment) \\
       Study 1  & Experiments 10-19 & Effect of perception failures \\
       Study 2  & Experiments 20-29 & Effect of driving policy failures \\
       Study 3  & Experiments 30-39 & Impact of late detections \\
       Study 4  & Experiments 40-49 & Impact of accident rate after late \\
                &                   & detection \\
       Study 5  & Experiments 50-59 & Combined effect of perception \\
                &                   & and driving policy failures
        \\ \hline
    \end{tabular}
    \caption{Overview of studies and experiments in Section~IV}
    \label{tab:studyoverview}
\end{table}

\subsection{Study 1: Effect of perception failures in different braking states on system safety }

In this study, we look at the impact of perception failures on system safety. Intuitively, we would expect that perception failures in different road conditions will affect the system safety differently. The model allows us to confirm such a conjecture and also to assess the magnitude of the difference. 
This study is conducted under several modeling assumptions:
\begin{itemize}
    \item The driving policy is assumed perfect, conditional on hazard being detected immediately upon entering a braking state.
    \item We also assume that an accident cannot occur during the period of overlooking the hazard and that all  hazards are eventually detected, following a short delay.
    \item Finally, if a hazard is overlooked, even for a short period of time, detecting it with some delay, may be too late for the adequate response to be enacted, thus the hazard may escalate to an accident.
\end{itemize}

\begin{figure}[t!]    
     \centering
     \begin{subfigure}[b]{0.48\linewidth}
         \centering
         \includegraphics[width=\linewidth]{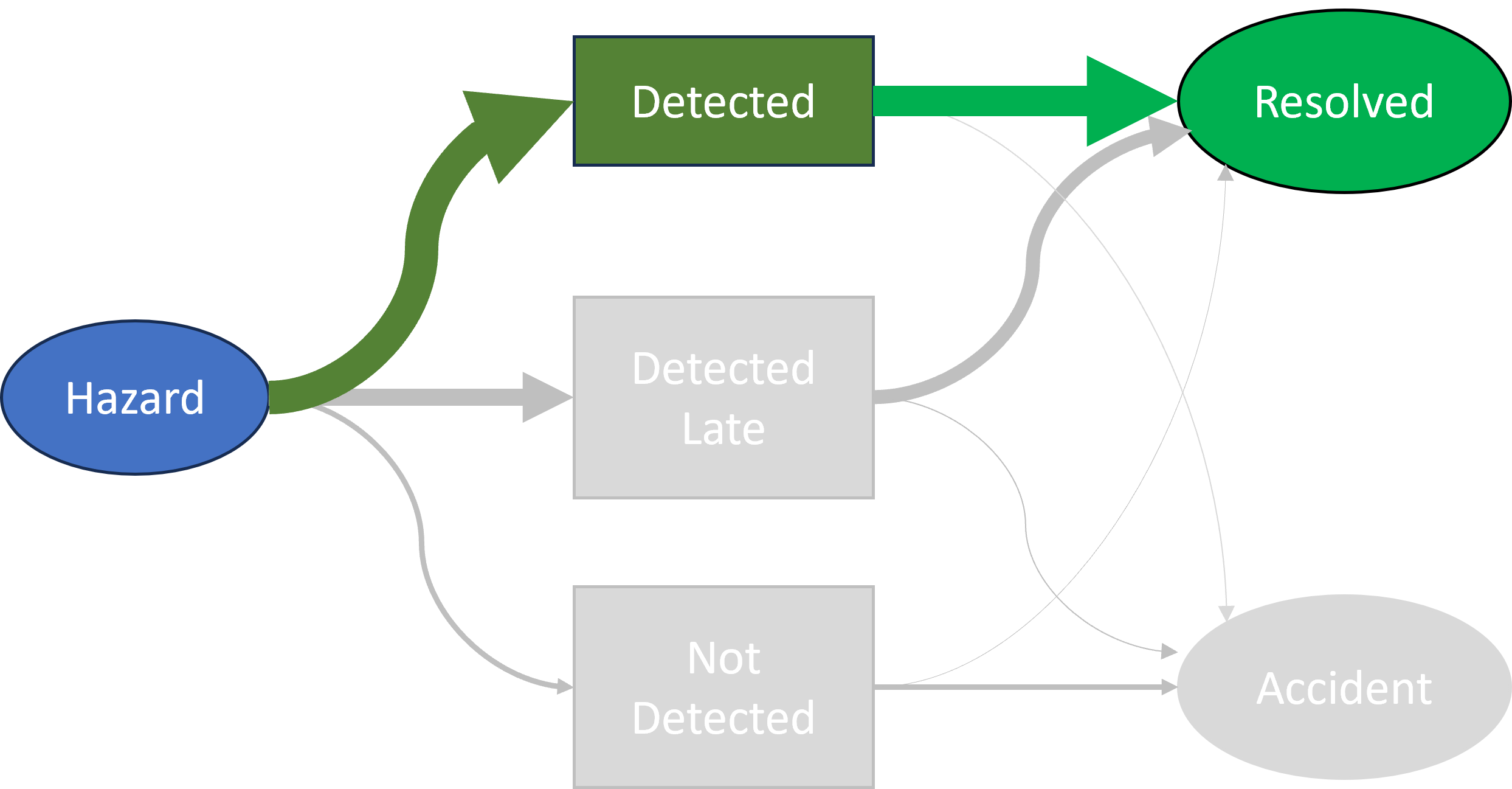}
         \caption{Experiment 1   \newline - Perfect perception}
         \label{fig:Exp10}
     \end{subfigure}
     \hfill
     \begin{subfigure}[b]{0.48\linewidth}
         \centering
         \includegraphics[width=\linewidth]{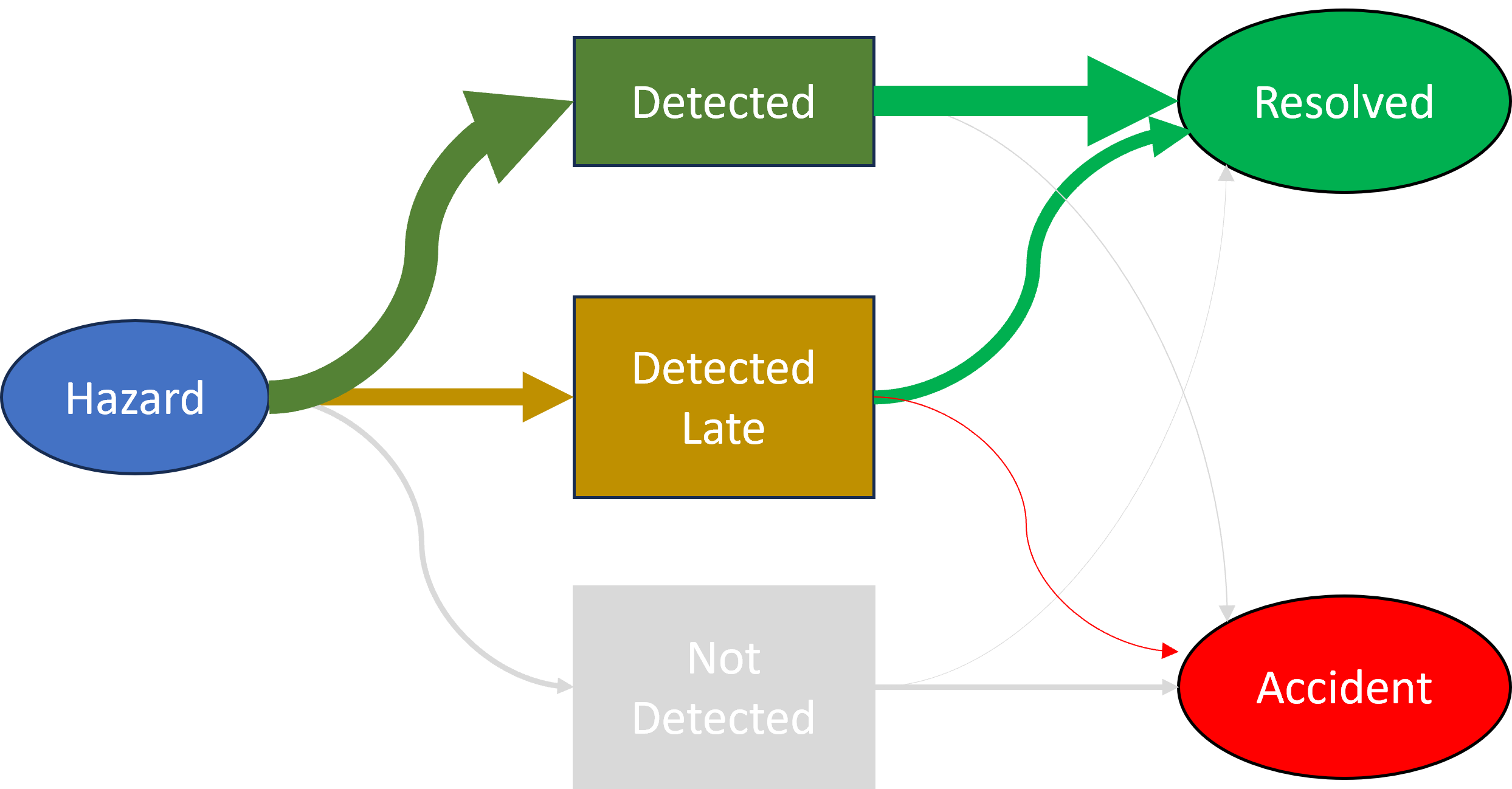}
         \caption{Study 1: Possible late detection}
         \label{fig:Exp1x}
     \end{subfigure}
     \caption{Possible state transitions for Study1}
     \vspace{-0.4cm}
\end{figure}

\begin{figure}[b!]
    \centering    
    \vspace{-0.4cm}
    \includegraphics[width=\linewidth]{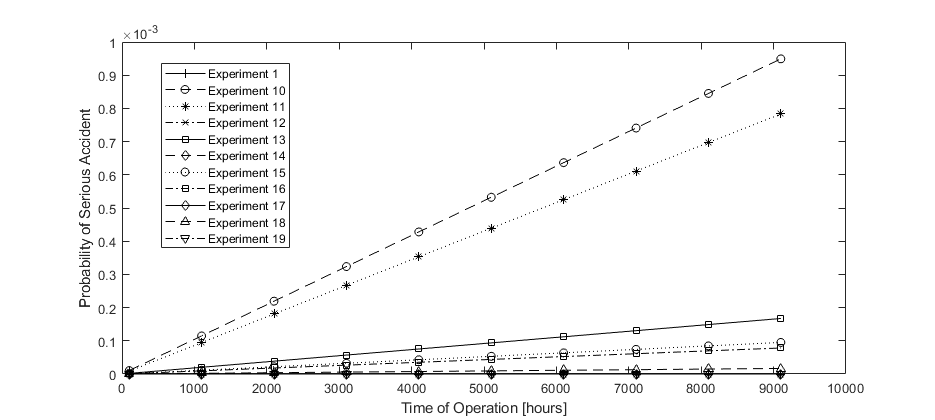}
    \caption{Probability of “serious accident” over a mission of different length for Study 1 (Experiments 10-19). Since Experiments 12, 14, 17 and 19 can only lead to a minor accident, the curves that represent these four experiments in the figure overlap with the curve of the "baseline" Experiment 1 (with no  serious accidents).}
    \label{fig:experiment1resultsserious}
\end{figure}

For the experiments we vary the probability of overlooking a hazard upon entering a hazardous state: i) Experiment 1 models a ``baseline'' scenario, in which the perception system works perfectly, hence a hazard will never be overlooked (see Figure~\ref{fig:Exp10}); ii) Experiment 11 – 14 are constructed under the assumption that a perception failure (overlooking a hazard) can occur in \textit{only one} of the hazardous states (F\_F\_B, F\_S\_B, IS\_F\_B, IS\_S\_B), while the perception is perfect in all other three hazardous states; iii) Experiment 10 models the situation when a perception failure can occur in all 4 hazardous states (see Figure~\ref{fig:Exp1x}). Likewise, in Experiments 16-19 the perception can fail in only one of the braking states, while with Experiment 15 perception failures can occur in all braking states. 
The probability of perception failure is assumed low, $10^{-3}$ and $10^{-4}$, for Experiment 10 - 14 and Experiment 15 - 19, respectively. Both, $10^{-3}$ and $10^{-4}$, are likely very optimistic, according to popular image classification benchmarks.


\begin{figure}[b!]
    \centering    
    \vspace{-0.2cm}
   \includegraphics[width=\linewidth]{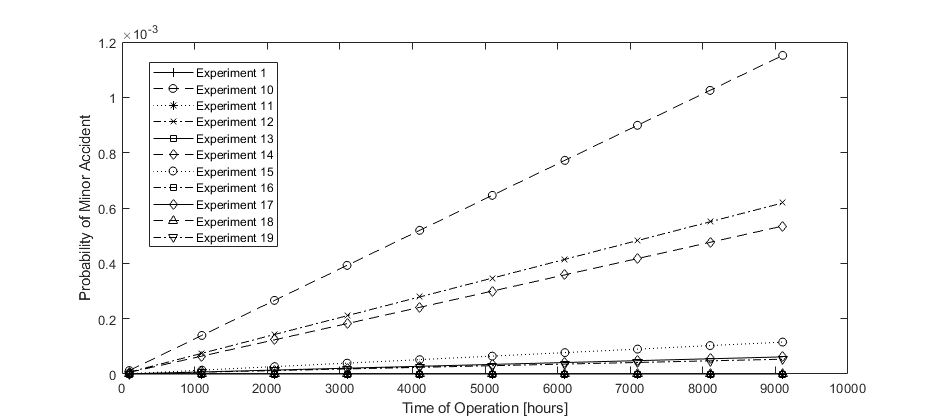}
    \caption{Probability of “minor accident” over a mission, as a function of mission duration, for Study 1 (Experiments 10-19). Since Experiments 11, 13, 16 and 18 can only lead to a serious accident, their curves in this figure overlap with the "baseline" Experiment 1 (no minor accidents).}
    \label{fig:experiment1resultsminor}    
\end{figure}

Solving the model for these different experiments leads to the results shown in Figure~\ref{fig:experiment1resultsserious} and Figure~\ref{fig:experiment1resultsminor}.
A closer inspection of the plots reveals that the ``base line'' Experiment~1 does not lead to accidents at all. This is expected as neither the perception system nor the driving policy would ever fail.\footnote{We recall that we are modeling the combined effects of road hazards and failures of AV functions. Thus we omit from the modeling those accidents that even perfect automation could not avoid. This needs to be taken into account when comparing the simulation results with accident statistics, which may well include this class of accidents.}  
Experiments 11, 13, 16 and 18 lead to serious accidents only, while Experiments 12, 14, 17 and 19 to minor accidents only. Experiment 10 and 15 may end up with either a serious or a minor accident. 
The observations are not surprising. Experiments 11, 13, 16 and 18 would have the perception system fail to detect a hazard upon entry for one of the fast braking states (X\_F\_B), which can eventually lead to a serious accident. Likewise, Experiments 12, 14, 17 and 19 would have the perception system fail to detect a hazard upon entry for one of the slow braking states (X\_S\_B), which can eventually lead to a minor accident. The effect of the probability of perception failure ($10^{-3}$ and $10^{-4}$, respectively) affect significantly the probabilities of accidents (serious or minor). Experiments 10 - 14 (with probability of perception failure set to $10^{-3}$) lead to probabilities of accidents, which are visibly greater than the probabilities of accident with Experiments 15 - 19 (with probability of perception failure set to ($10^{-4}$). While the ordering is not surprising the model provides an insight about the magnitude of the impact of the probability of perception failure on the probability of accident. Finally, Experiment 10 and 15 are such that perception failures can occur in all braking states, which in turn can lead to either a serious or a minor accident. The plots representing these two Experiments show a clear ordering of the probabilities of accident. After 9100 hours of operation, the probabilities of accident with Experiment 10 is $10^{-3}$ and $1.2 \times 10^{-3}$ for the serious and minor accidents, respectively. The  probabilities of accident for Experiment 15 are an order of magnitude smaller.

This study exemplifies how the impact on system safety could vary with the location of the perception failures: as evident from Figure~\ref{fig:experiment1resultsserious} the same probability of perception failure of $10^{-3}$ (or $10^{-4}$) in a fast state will affect system safety more significantly if it occurs during free driving (Experiment 11 or Experiment 16) than if the perception failure occurred in an intersection state (Experiment 13 or Experiment 18). In contrast for slow driving the impact on safety is higher if the error occurs in an intersection.

If an automotive developer were to derive recommendations from this study, the practical implication would be that investment on improving a perception system should be directed with top priority to improving the perception function in free, fast driving (e.g., improving the algorithms of object recognition on motorways), then on fast driving in intersections, followed by improvements in intersection at low speed and finally in slow free driving. Such focused improvements of perception could be achieved using different techniques, e.g., including using dedicated perception channels to deal with each of the hazardous states, rather than one single monolithic perception component, which may be more difficult to implement in such a way as to allow focused improvements in specific operating conditions.

\begin{figure}[t!]
     \centering     
     \begin{subfigure}[b]{0.48\linewidth}
         \centering
         \includegraphics[width=\linewidth]{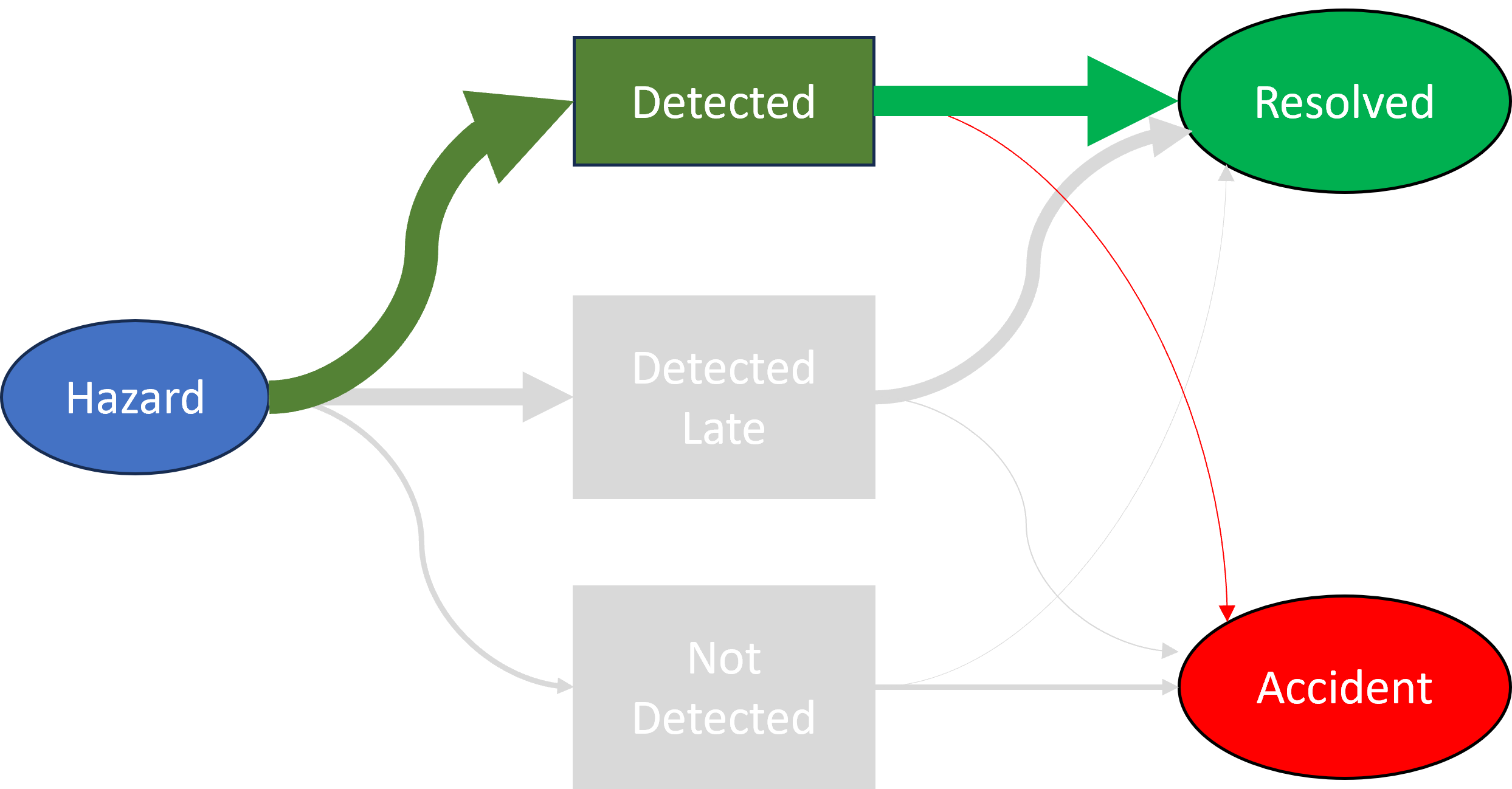}
         \caption{Study 2:   \newline Driving policy failure}
         \label{fig:study2}
     \end{subfigure}
     \hfill
     \begin{subfigure}[b]{0.48\linewidth}
         \centering
         \includegraphics[width=\linewidth]{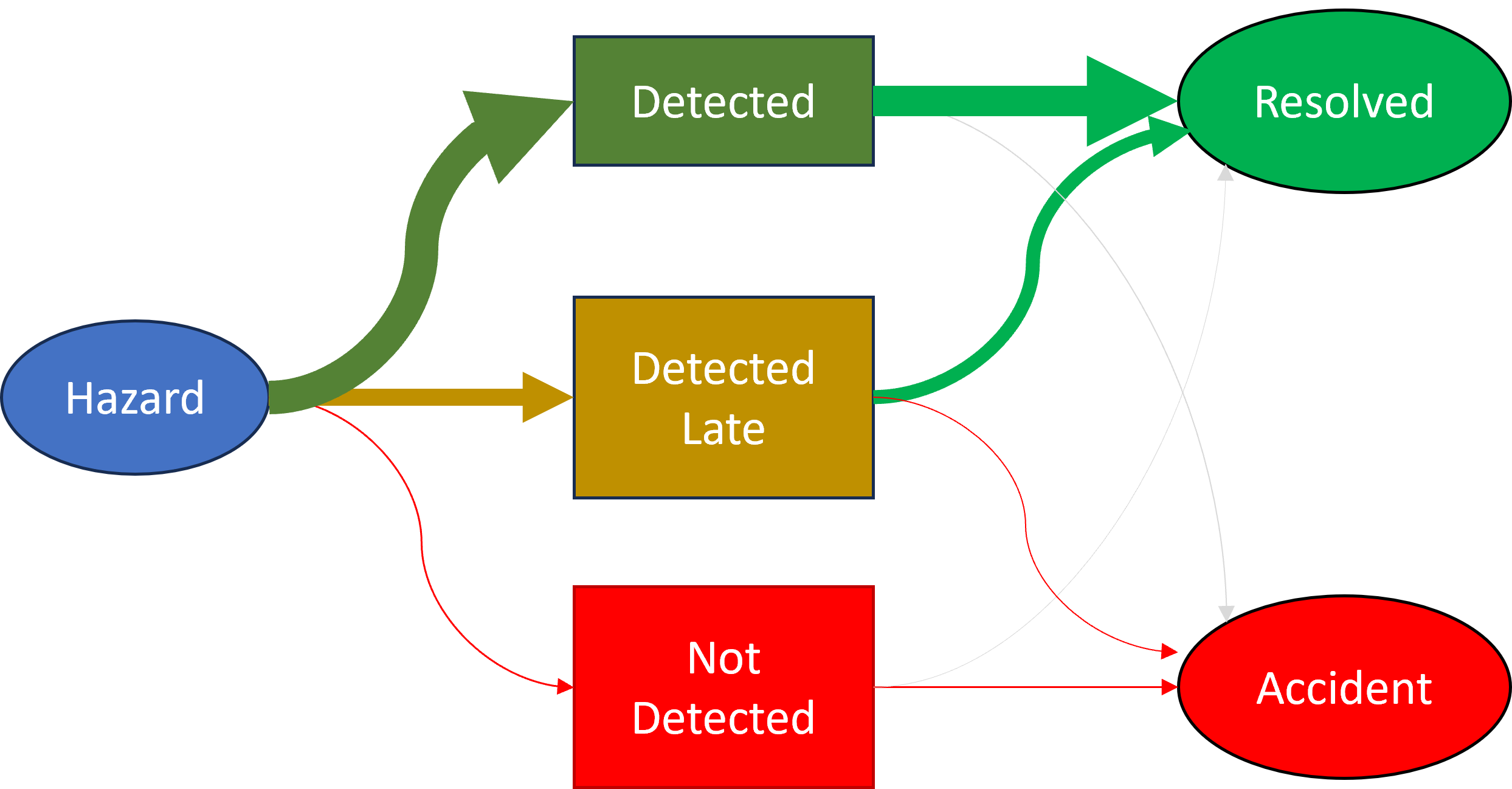}
         \caption{Study 3: Impact of late detection}
         \label{fig:Study3}
     \end{subfigure}
     \caption{Possible state transitions for Study 2 and Study 3} 
     \vspace{-0.4cm}
\end{figure}

\subsection{Study 2: Impact of driving policy reliability on AV safety }
In this study we look at the impact of driving policy’s reliability on system safety by varying the failure rate of the driving policy in the hazardous states. The study was again conducted with a set of different experiments.
Here, the only parameters which are assigned non-zero values are the rates of failure of the driving policy when the road hazards are properly and timely detected, which means that the perception works flawlessly, but the driving behavior in response to detected hazards may be erroneous and may lead to accidents (see Figure~\ref{fig:study2}).
For Experiments 21 - 24 and 26-29 the driving policy can fail only in one of the hazardous states, but is perfect in the others hazardous states. In Experiments 20 and 25 the driving policy can fail in any of the hazardous states. For Experiments 20 - 24, the rates of failure are set to $10^{-6}$, while for Experiments 25 - 29 the rates of failure are set to $10^{-5}$, thus allowing us to study the impact of the rate of failure of the driving policy in different hazardous states on system safety. 
\par
\begin{figure}[t!]
    \centering    
    \vspace{-0.2cm}
    \includegraphics[width=\linewidth]{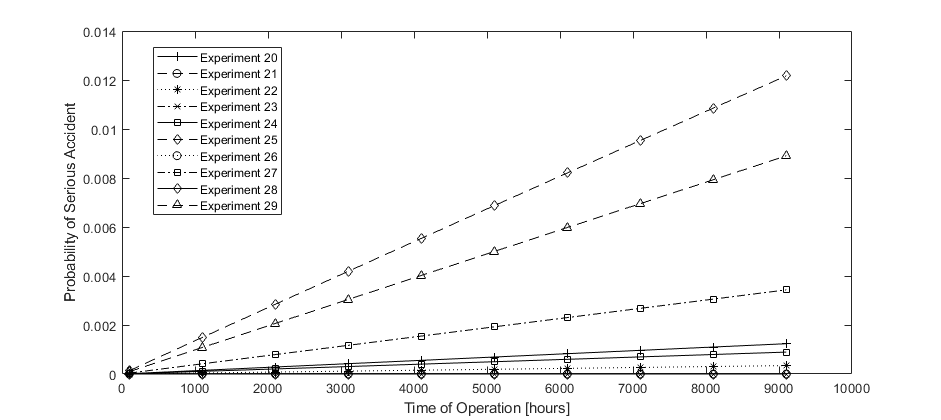}
    \caption{Probability of “serious accident” over a mission, as a function of mission duration, for Study 2 (Experiments 21-29). The curves of Experiments 22, 24, 27 and 29 appear as straight horizontal overlapping lines at the bottom of the figure  as these experiments do not lead to serious accidents.}
    \label{fig:experiment2resultsserious}  
    \vspace{-0.4cm}
\end{figure}
\begin{figure}[b!]
    \centering     
    \vspace{-0.4cm}
    \includegraphics[width=\linewidth]{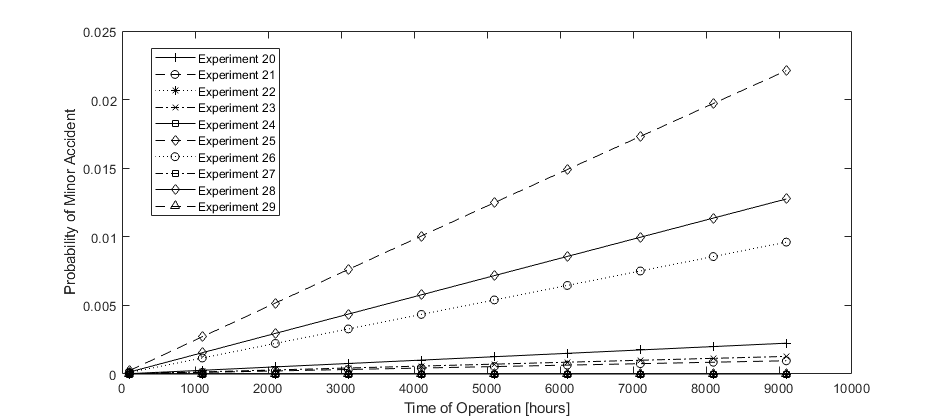}
    \caption{Probability of “minor accident” over a mission, as a function of mission duration, for Study 2 (Experiments 21-29). The curves representing Experiment 21, 23, 26 and 28 appear as overlapping straight lines at the bottom of the figure as these experiments do not lead to minor accidents.}
    \label{fig:experiment2resultsminor}        
\end{figure}

The results from solving the experiments in this group are shown in Figure \ref{fig:experiment2resultsserious} and Figure \ref{fig:experiment2resultsminor}

As with Study 1, one can see that the experiments are split into two groups. Experiments 21, 23, 26 and 28 may lead to a serious accident only, while Experiments 22, 24, 27 and 29 may lead to a minor accident only. The split is expected and is a consequence of the location of the driving policy failure: fast states for Experiment 21, 23, 26 and 18 (which can escalate to a serious accident, only) and slow for Experiments 22, 24, 27 and 29 (which can escalate to a minor accident only), respectively. 

One can see in Figure~\ref{fig:experiment2resultsserious} and Figure~\ref{fig:experiment2resultsminor} that the probabilities of accident in these experiments are considerably greater than the respective probabilities in the previous Study 1: after 9100 hours of operation the probability of accident recorded in Experiment 25 (the worst of all experiments in this study) reaches 0.012 for serious accidents and slightly exceeds 0.02 for minor accidents. These levels of probability of an accident are two orders of magnitude higher than the levels recorded for the worst experiment in Study 1. Although the nature of the parameters in Study 1 and in this study is different - probabilities of perception failure vs. rate of failure of the driving policy - the observation seems indicative of how different the impact of these two model parameters on system safety is: even very small rates of failure of the driving policy ($10^{-6}$ to $10^{-5}$) lead to significantly greater probability of accident than does a perception system with probability of failure in the range of $10^{-3}$ to $10^{-4}$.\par 
The probability of minor accident in this study is more than twice greater than the probability of serious accidents (see Experiments 20 and 25, respectively). In study 1 the probabilities of minor and serious accidents (Experiment 10 and 15) were very close (around $10^{-3}$ in both cases).

Finally, comparing the two groups of experiments in this study (Experiments 20 - 24 and 25 - 29) conducted with rates of failure of the driving policy, which are an order of magnitude apart, we observe that the probabilities of accident increase/decrease according to the rate of failure of the driving policy: the probability of minor accident after 9100 hours of operations for Experiments 20 and 25 is slightly about 0.001 and slightly over 0.012, respectively (i.e., an order of magnitude apart). This observation provides an additional insight about the impact of the different modeling parameters (i.e., the reliability of the components they model) on the probability of accident. 

This analysis, illustrated in Figure~\ref{fig:experiment2resultsserious} and Figure~\ref{fig:experiment2resultsminor}, confirms the observations that we made in~\cite{Buerkle2022} that AV safety is very sensitive to the reliability of its driving policy. Based on the present model, even a very low probability of driving policy failure will imply modest safety, especially if the rate of failures of the driving policy of $10^{-5}$: the worst experiment in this study (Experiment 25, where failures of the driving policy can occur in all hazardous states) after 9100 hours of operation will lead to an accident with probability exceeding 3\% (either serious or minor), which seems to indicate poor safety. For the more reliable driving policy used in the study (with rate of failure set to $10^{-6}$) the probability of accident (Experiment 20) after 9100 hours is well below 1\%.

Demonstrating that a driving policy will have a probability of failure of $10^{-6}$ may be very challenging, due to the complexity of such a component.
Therefore, various authors propose to use a safety monitor to assure correct behavior of the driving policy ~\cite{shalev2017formal,oboril2021rss+,nister2019introduction}. 
Recall that we included in the AV driving policy both the AV planning and safety monitors. So the used failure rate reflects failure of the combination of driving policy and safety monitor.

\subsection{Study 3: Impact on safety of late hazard detections}

Now we scrutinize the impact of overlooking the road hazard for a short period of time upon entering hazardous states. The study is conducted under the following assumptions (see Figure~\ref{fig:Study3}):
\begin{itemize}
    \item Driving policy rates of failure in hazardous states are set as follows: i) conditional on timely and correct hazard detection the rates are set to  $10^{-5}$ (Experiments 30 - 34) or $10^{-6}$ (Experiments 35 - 39), respectively. ii) Conditional on late hazard detection, the rates are set to $2 \times 10^{-5}$ (Experiments 30 - 34) and $2 \times 10^{-6}$ (Experiments 35 - 39), respectively, i.e., to values which are twice greater than the values assigned to the rate conditional on timely and correct hazard detection. Experiments 30 and 35 allow failures of the driving policy in all hazardous states, while the other experiments included in this study allow failures of the driving policy in one hazardous state only. 
    \item The rates of accident after late hazard detection (i.e., an accident occurs \textit{before} the hazard is eventually detected albeit with a small delay) are set to  $10^{-5}$ for all hazardous states and all experiments. 
    \item The probability of failure of the perception system is set to $10^{-4}$ for all hazardous states. Experiment 30 and 35 allow perception failures in all hazardous states, while the other experiments included in the study allow perception failure in only one of the hazardous states.
    \item The delays in detecting hazards are defined in Table III. 
\end{itemize}

\begin{figure}[t!]
    \centering    
    \includegraphics[width=\linewidth]{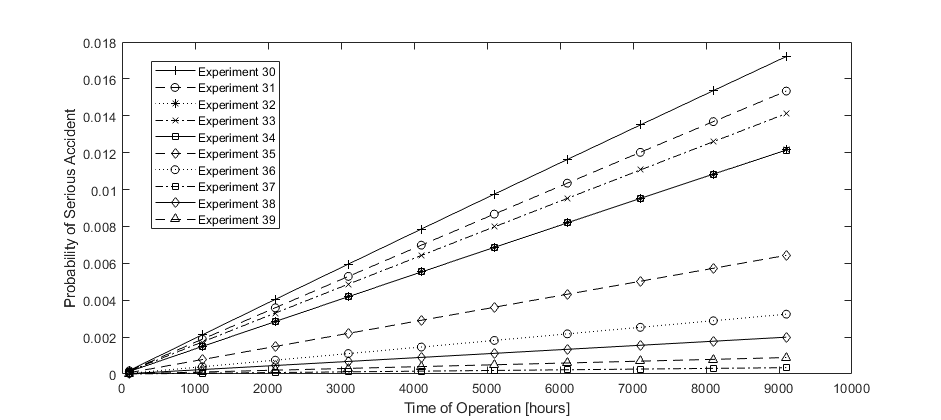}
    \caption{Probability of “serious accident” over a mission, as a function of mission duration, for Study 3 (Experiments 30-39)}
    \label{fig:experiment3resultsserious}
    \vspace{-0.4cm}
\end{figure}

\begin{figure}[b!]
    \centering  
    \vspace{-0.4cm}
    \includegraphics[width=\linewidth]{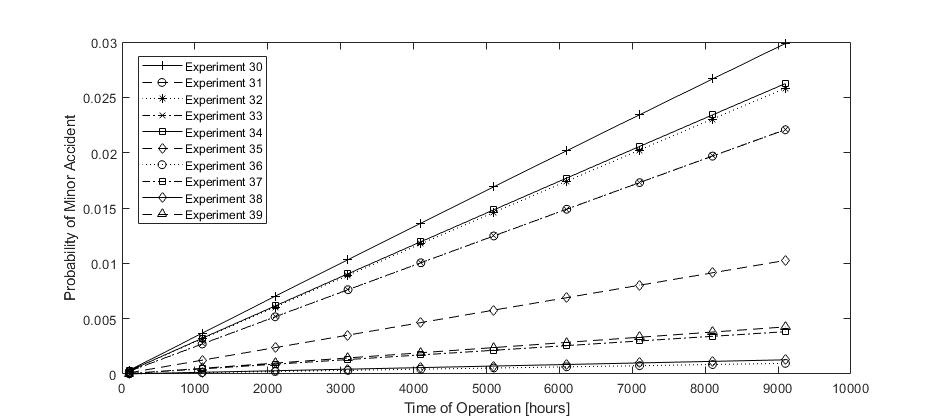}
    \caption{Probability of “minor accident” over a mission, as a function of mission duration, for Study 3 (Experiments 30-39)}
    \label{fig:experiment3resultsminor}    
\end{figure}

The results from solving the experiments included in the study are presented in Figure \ref{fig:experiment3resultsserious} and Figure \ref{fig:experiment3resultsminor}. Clearly, the difference between the experiments is in the reliability of the perception system – it varies between the hazardous states, in terms of rate of failure of the driving policy  and the probability of accident at the end of the short period of overlooking a hazard in one of the hazardous states. 

The results show a more nuanced picture of how the model parameters (which model the respective components' reliabilities) affect system safety. There is a clear separation between the two groups of experiments (Experiment 30 - 34 and Experiment 35 - 39, respectively) for which some of the parameters differ by an order of magnitude, as explained above. The probability of accident, however, now differs by a factor of 3 between the comparable experiments: after 9100 hours of operation, Experiment 30 will have a serious accident with probability of 0.018 and a minor accident with probability 0.03. For Experiment 35 these probabilities are 0.006 and 0.01 for the serious and minor accidents, respectively. The study, thus, provides  an insight about the effect of model parameters on the chosen measure of safety, which is different (less-well pronounced) than for the previous two studies, in which the safety measure was more sensitive to the change of model parameter values.



\begin{figure}[t!]
     \centering     
     \begin{subfigure}[b]{0.48\linewidth}
         \centering
         \includegraphics[width=\linewidth]{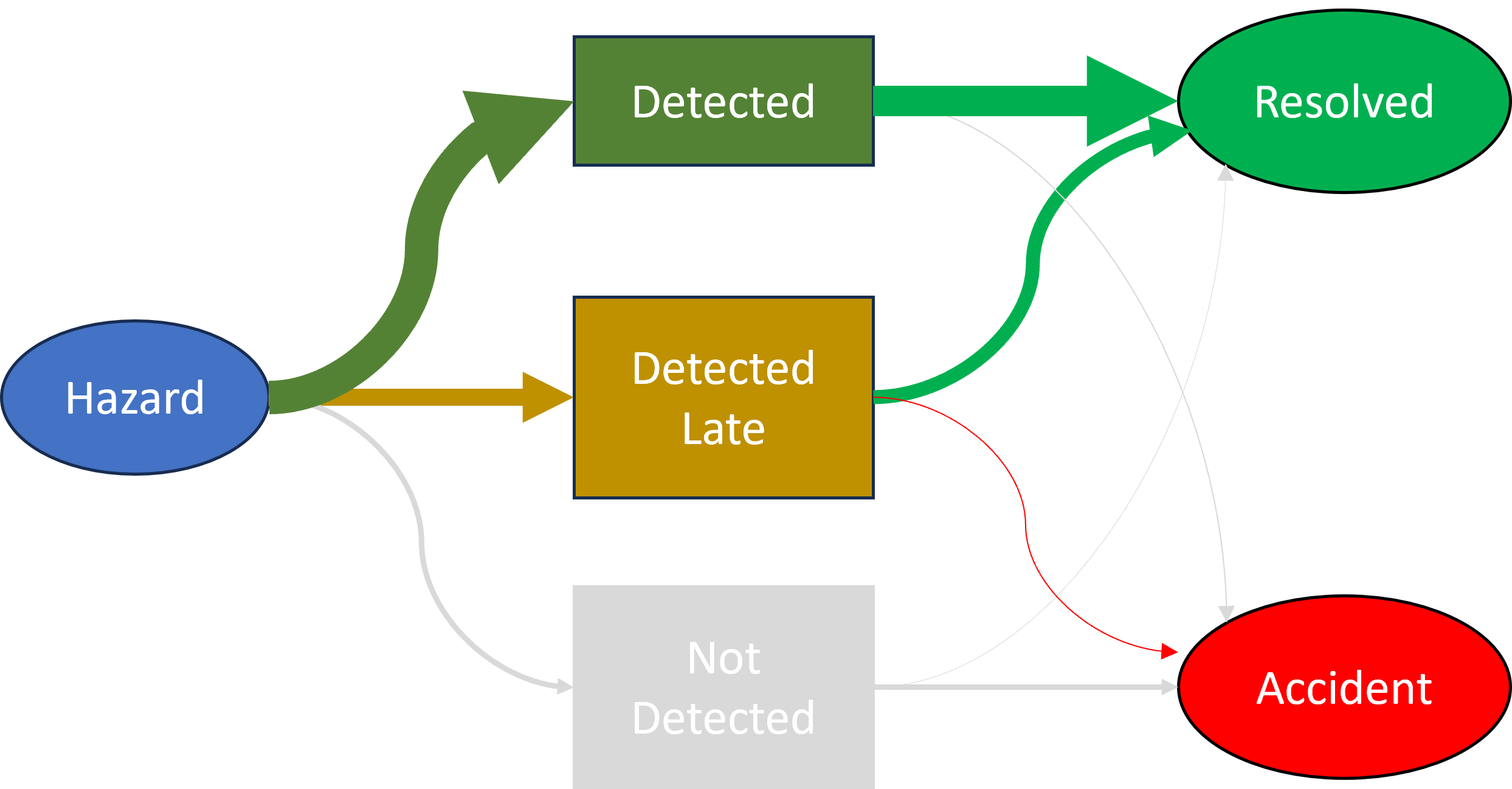}
         \caption{Study 4: Impact of accident rate after late detection~\\}
         \label{fig:study4}
     \end{subfigure}
     \hfill
     \begin{subfigure}[b]{0.48\linewidth}
         \centering
         \includegraphics[width=\linewidth]{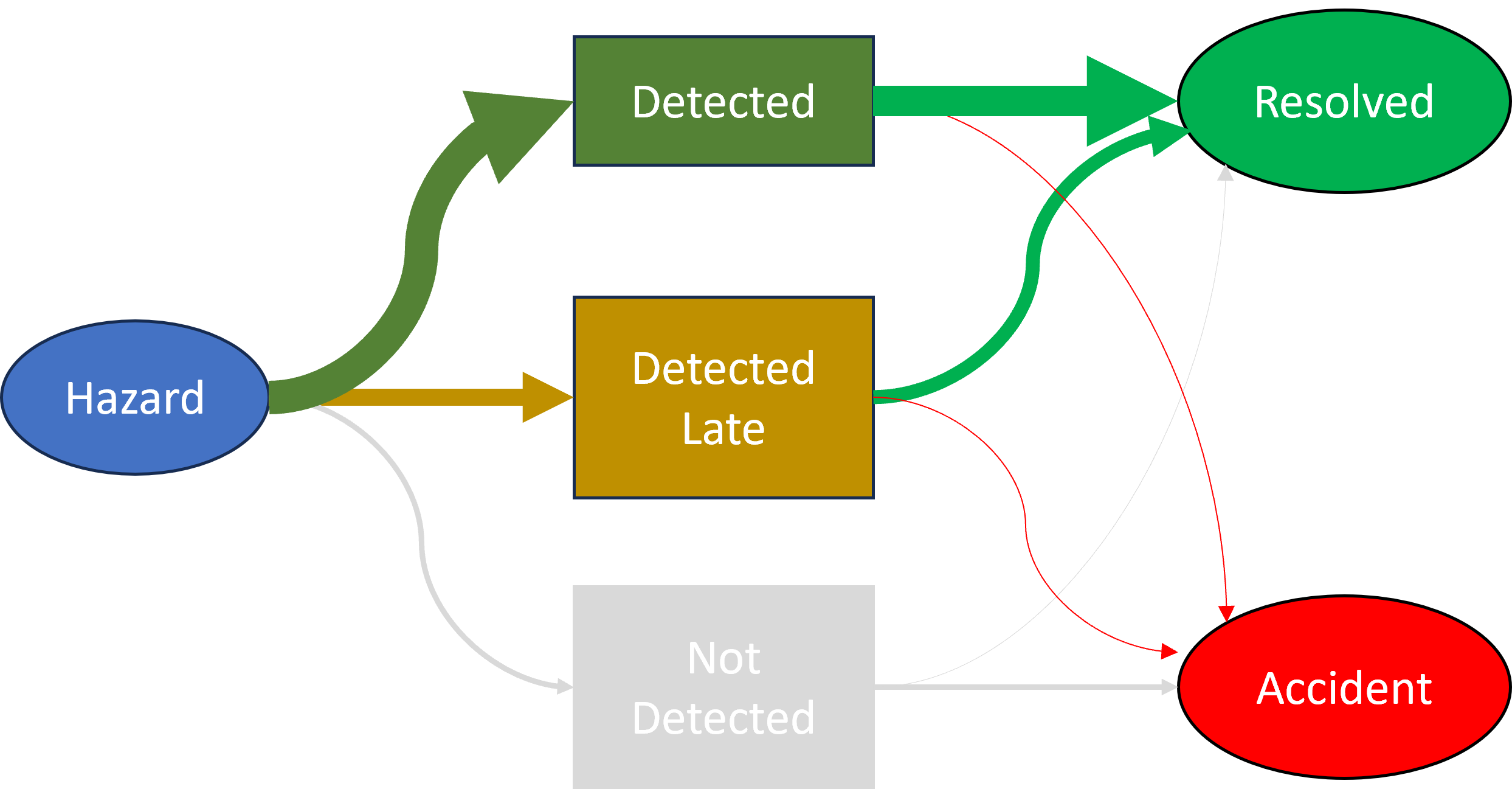}
         \caption{Study 5: Combined effect of perception and driving policy failures }
         \label{fig:Study5}
     \end{subfigure}
     \caption{Possible state transitions for Study 4 and Study 5} 
     \vspace{-0.4cm}
\end{figure}

\subsection {Study 4: Impact on AV safety of simultaneous failures of driving policy and perception system }

This study is somewhat similar to Study 1 in the sense that the rate of failure of the driving policy, conditional on timely and correct detection of hazards is set to 0 (i.e., the policy is perfect conditional on correctly working perception). We scrutinize the effect of failure of the driving policy conditional on late detection and set the modeling parameters as follows:
\begin{itemize}
    \item The driving policy is perfect conditional on timely and correctly detecting a hazard. The driving policy can fail conditional on perception failure, however. The conditional rates of failure conditional on late hazard detection are set to $10^{-4}$ (Experiment 40 - 44) and to $10^{-3}$ (Experiment 45 - 49). Experiments 40 and 45 allow failures of the driving policy in all hazardous states, while the other experiments included in the study allow failures of the driving policy in only one of the hazardous states.  
    \item The probability of accident following a late detection of the hazard is set to 0;
    \item Perception failures can occur in all experiments in the study and for all hazardous states with the same probability of $10^{-4}$, a value consistent with the previous studies and quite small given the measured reliability of real perception systems used in AV. 
\end{itemize}

\begin{figure}[b!]
    \centering  
    \vspace{-0.4cm}
    \includegraphics[width=\linewidth]{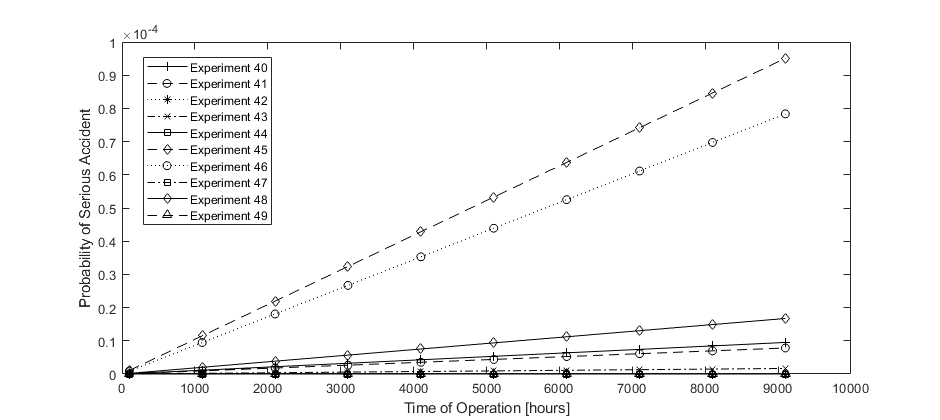}
    \caption{Probability of “serious accident” over a mission, as a function of mission duration, for Study 4 (Experiments 41-45)}
    \label{fig:experiment4resultserious}        
\end{figure}

A simplified state transition diagram for this study is illustrated in Figure~\ref{fig:study4}. 

\begin{figure}[t!]
    \centering     
    \vspace{-0.4cm}
    \includegraphics[width=\linewidth]{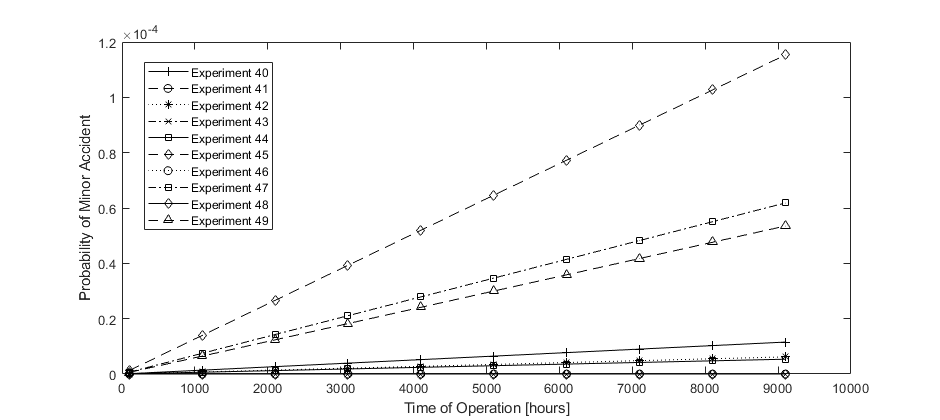}
    \caption{Probability of “minor accident” over a mission, as a function of mission duration, for Study 4 (Experiments 41-45)}
    \label{fig:experiment4resultsminor}        
    \vspace{-0.4cm}
\end{figure}

The corresponding results, depicted in Figure~\ref{fig:experiment4resultserious} and Figure~\ref{fig:experiment4resultsminor}, are in line with our expectations: Experiment 45 is the worst in terms of both – probabilities of serious and minor accidents, although the probabilities of accidents at the end of the observation of 9100 hours is quite small – much smaller than $1\%$ (in the order of 0.01\% for both serious and minor accidents). While the impact of the location of imperfection is clear to see ( consistent with the observations in the previous studies), the magnitude of the difference is somewhat surprising. For instance, the difference between the probabilities of serious accidents recorded for Experiment 46 and 48 (Figure~\ref{fig:experiment4resultserious}) is much greater than the difference of the same probability between 48 and 40, despite the significant difference of driving policy failure rate between Experiment 48 and 40. This pattern is different from the grouping of experiments shown in Figure~\ref{fig:experiment4resultsminor} where Experiment 45, 47 and 49 form a distinctly different group from the other experiments which can produce minor accidents (Experiment 40, 42 and 44).

\subsection{Study 5: Combined effect of perception failures \& failures of driving policy}	

The experiments included in Study 5 are chosen to allow a more systematic exploration of different failure scenarios that can occur in operation, thus offering a more fine-grained insight about the combined effect of failures of perception (i.e., overlooking a hazard upon entering a hazardous state), and of the ability to recover from a delayed detection of a hazardous state after a short period of overlooking a hazard, combined with a driving policy of limited reliability. 

Intuitively, we would expect that the combined effect of several failures will be worse than the effect due to a single failure and the study provides an insight about the magnitude of the difference. The model parameters are assigned values as follows:
\begin{itemize}
    \item The driving policy rate of failure conditional on timely and correct detection of a hazard varies between $10^{-6}$ (Experiment 50 - 54) and $10^{-5}$ (for Experiment 55 - 59). The rate of failure of the driving policy conditional on late hazard detection varies between $10^{-5}$ and $10^{-3}$ as follows: it is set to $10^{-5}$ for Experiment 50, 52 and 59, to $10^{-4}$ for Experiment 51, 53 and 56, to $2 \times 10^{-4}$ for Experiment 57, to $5 \times 10^{-4}$ for Experiment 58, and to $10^{-3}$ for Experiment 54 and 55. The assignment of parameter values followed the rule that the  failure rate of the driving policy conditional on perception failure should be no smaller than the rate of failure conditional on timely and correct hazard detection.
    \item The probability of accident following a late detection of the hazard is set to 0, e.g., no accident can occur before an overlooked hazard is (eventually) detected albeit with some delay. 
    \item The probability of perception failure varies between $10^{-6}$ and $10^{-4}$ as follows: it is set to $10^{-6}$ for Experiment 50 and 51, to $10^{-5}$ for Experiment 52 - 54, to $10^{-4}$ for Experiment 55 to 59. Perception failures may occur in any hazardous state for all 10 experiments included in the study. 
\end{itemize}


\begin{figure}[t!]
    \centering  
    \includegraphics[width=\linewidth]{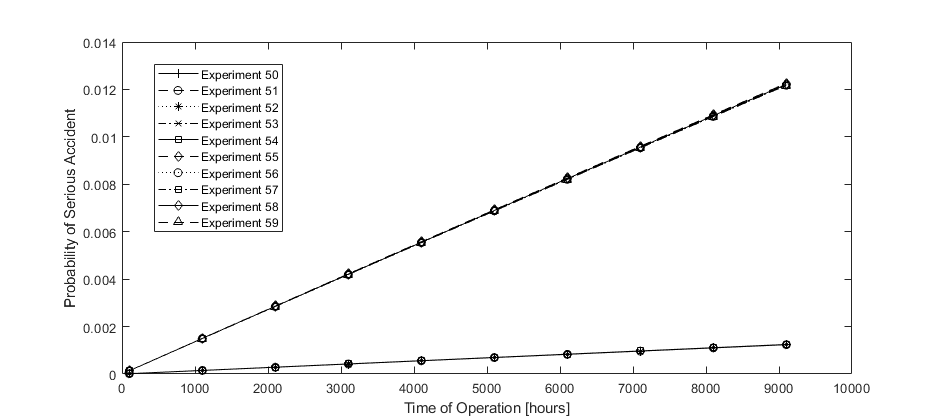}
    \caption{Probability of “serious accident” over a mission, as a function of mission duration, for Study 5 (Experiments 50-59). Note: Experiments 50-54 provide similar results (lower line), and 55-59 are also similar (upper line)}
    \label{fig:experiment5resultserious}    
    \vspace{-0.4cm}
\end{figure}

\begin{figure}[b!]
    \centering      
    \vspace{-0.4cm}
    \includegraphics[width=\linewidth]{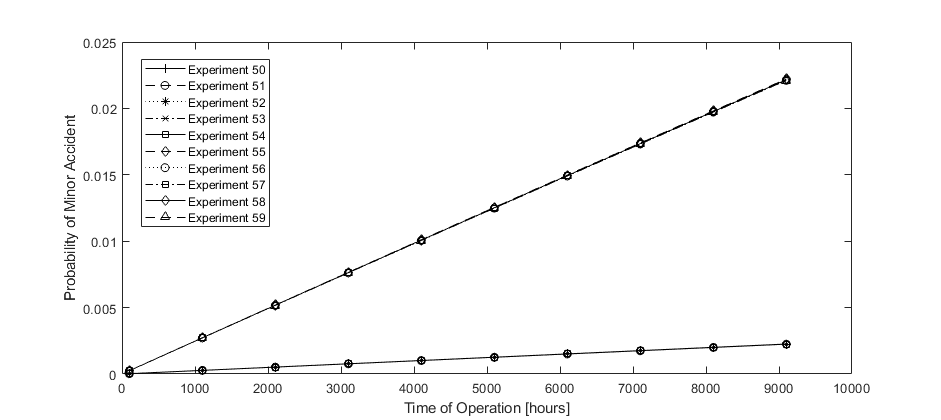}
    \caption{Probability of “minor accident” over a mission, as a function of mission duration, for Study 5 (Experiments 50-59). Note: Experiments 50-54 provide similar results (lower line), and 55-59 are also similar (upper line)}
    \label{fig:experiment5resultsminor}    
\end{figure}

The results from the study are shown in Figure~\ref{fig:experiment5resultserious} and Figure~\ref{fig:experiment5resultsminor}. Quite clearly, there are two distinct groups of almost identical plots: i) group 1 includes Experiment 50 - 54, ii) group 2 - Experiment 55 - 59. It seems clear that the dominant parameter, which was assigned the same value in Experiment 50 - 55 and Experiment 55 - 59, respectively, is the rate of failure, conditional on detecting a hazard on time and correctly. The variation of the other parameters has a negligible impact on system safety. This observation seems quite significant, as it reiterates the dominant role  the rate of failure of the driving policy (including of safety monitors) has on system safety. The limited impact of the other parameters on system safety could be easily explained: in those cases when perception works correctly, the risk of having an accident is dependent entirely on the rate of failure of the driving policy conditional on hazard being detected correctly and on time. The higher rate of failure of the driving policy conditional on perception failure is mitigated by the small probability of perception failures. For some of the experiments the ratio between the conditional rates of failure given perception failure and correct and timely hazard detection is 2-3 orders of magnitude, e.g., for Experiment 54 and 55 the rate conditional on perception failure is set to 0.001 against $10^{-6}$ or $10^{-5}$ assigned to the conditional rate of failure given correct and timely hazard detection. Clearly, for the increased rate of failure of the driving policy conditional on delayed hazard detection to manifest itself and to cause an accident, a "simultaneous" failure of the perception system is needed. In other words, an accident will only occur if both the perception and the driving policy fail to process correctly the \textit{same instance} of a hazardous situation. Such simultaneous failures are unlikely and significantly less likely (given the chosen model parameterization, of course) than the occurrence of correct hazard detection combined with a failure of the driving policy. This particular observation with the model is an  example of the value of "defense in depth", a principle widely used in safety engineering.

\section{Discussion}
We have extended the model-based approach to safety assessment of AV, presented earlier \cite{Buerkle2022} based on probabilistic modeling. The modeling approach allows one to spell out phenomena that affect AV safety and then explore in quantitative fashion the impact of model parameters on AV safety. 

We demonstrated that some of the model parameters can be estimated using naturalistic driving data that are already being collected by automotive companies. For some other model parameters, no suitable data existed, and we had to apply sensitivity analysis varying their values within plausible ranges (also exploring ranges of parameter values beyond the currently achievable levels of reliability of perception and driving policy functions). It turned out that, with these ranges of parameter values, some of the model parameters had little to no impact on AV safety, e.g., the delay in perceiving initially overlooked hazards, and duration of false hazards have negligible effect compared to the rate of failure of driving policy. Other parameters, however, e.g., the probability of overlooking a hazard, given that the hazard exists, affect AV safety significantly. The results obtained with the model suggest that even with a very good perception system (probability of overlooking a hazard of $0.01\%$), the AV’s safety is modest – the probability of an accident of over a year (9100 hours non-interrupted driving) is non-negligible. While 9100 hours of driving at high speed means over 0.5 million km of driving (for an average velocity on the road of say 60 km/h), these figures still compare unfavorably with human drivers’ average accident rates (of the order of hundreds of millions kilometers between fatalities and several millions between injuries \cite{Transport2021}). If this safety level is unacceptable, developers need to improve some of the parameters under their control, by improving: i) the AV perception, which affects the critical model parameter, the probability of overlooking a hazard, which we already assumed very good; and/or ii) the safety mechanisms, which affect the probabilities of transition to the accident state.

The role of sensitivity analysis, which we demonstrated throughout all 5 studies, is noteworthy. It allows one to identify those parameters that affect AV safety significantly within plausible ranges and concentrate on estimating them accurately or even conservatively. Parameters that have little/no impact on model behavior (e.g., in our case, the delay in perceiving hazards that are not identified immediately) require less estimation precision. Demonstrating that these parameters lie within a “ballpark” range for which models show adequate safety will be sufficient, which can save the effort and resources that would be needed for more accurate estimates.  
Another aspect worth mentioning is that the values of the various parameters are usually not constant over time. 
For instance, traffic densities, vehicle performance, driving behavior will evolve over the years.
However, in this work we assumed that the distributions of the various intervals, captured in the model by timed activities, do not change over time. This is clearly a simplification: the distributions (and their parameters, e.g., the rates of exponential distributions) typically will vary over time. The impact of this variation will pose no technical problems – the theory of non-homogeneous Markov processes is well developed and should be easy to apply in our proposed model.

In terms of estimating the model parameters that may vary over time, we do not see any conceptual difficulty, either. For instance, the method adopted for collecting the datasets used to estimate the transition rates between the 8 states we used in the model seems applicable in circumstances where the parameters may vary. Thus, parameter estimation can be updated dynamically as necessary; if trends are noticed, the user of the model can try to extrapolate the trend so as to estimate the level of safety in a hypothetical, but expected, future. 
A related problem concerns variation of driving conditions in operation.
Operational Design Domain (ODD) captures the idea well but focuses on ways of capturing the differences between operation conditions (“modes of the environment”) within a given ODD. The dynamic aspects of how these operating conditions manifest themselves in operation, e.g., the rate of change of these operating conditions, is paid little attention but can be easily integrated by i) using the dynamic approach explained above, and ii) sub-models capturing different parameter values for different conditions within the chosen ODD. The extension that we have presented in this work illustrates how ODDs can be included in the safety analysis: our model operates with an ODD which consists of 8 different operating conditions – some hazardous, some not, including different speed of driving. 
In our judgment the approach can be scaled to a much larger number of operation conditions, e.g. 100, 1000 and tens of thousands. A difficulty with such large models will be in estimating the model parameters and can be done in the same manner as in this paper – relying on naturalistic datasets that might be available and using sensitivity analysis for those parameters for which no suitable data set exists. We do not envisage any technical difficulties in solving the resulting large stochastic model, as the particular technique and tool for solving the models have been used with very large models (of tens of thousands of states). 

Among the model parameters which we dealt with via sensitivity analysis, some are critical for the safety estimates we derive. 
For instance, the rate of accident conditional on a road hazard detected immediately upon its occurrence, or detected after a delay. Collecting data to estimate these rates does not represent a conceptual difficulty. Recording how many road hazards led to accidents and how many road hazards did not is all that is needed to estimate the respective model parameters. The complexity of this data collection will be no greater than collecting data on “near misses” with AV advocated by many authors, e.g. \cite{Johansson2022}, as useful in safety assessment. However, the rarer the events to be counted, the broader the confidence intervals around the estimates derived from these counts. Indeed, by the time enough experience has been accumulated that one could state within narrow confidence bounds the accident rate of an AV (assuming that the conditions of traffic in the future reproduce those in the past exactly), using the same data instead to feed parameters into our model would yield predictions with broader uncertainty intervals.
The value of detailed modeling approaches like ours is instead in preliminary assessments, allowing designers to allocate allowable failure probabilities between subsystems, or to assess likely safety levels in a different environment (e.g. different country, or a foreseeable future with different prevalence of AVs in traffic).

We mentioned in passing that discriminating between different types of accidents ("modes of catastrophic failure”) is outside the scope of this work. However, without trying to be comprehensive, we already offered initial advances in this direction by defining two types of accident “serious” and “minor”. These are attributed in the model to hazardous states with different velocity of vehicle ("fast" and "slow"). In practice, safety analysis typically discriminates between different accidents, e.g. \cite{NHTSA2022}. In such cases we could extend the approach further by modeling the road hazards in more details to allow for attributing the accidents to their respective causes. Again, the difficulties will be technical - in collecting data which will allow for parameter estimation – rather than conceptual in terms of capturing in the model the required additional level of detail.

Finally, the influence of hardware errors vs. software errors should be noted. The  impact of hardware failure rates to the perception system is low, as the hardware impact is typically managed by hardware development and production processes, which are captured in safety standards like ISO26262 \cite{ISO26262}. Additionally, mechanisms for modern perception machine learning architectures are available that can limit the hardware impact on system level perception failures \cite{Geissler2021}. Also, typically machine learning models are quite robust to hardware failures. As a result, hardware errors, which are to be expected in the ballpark of $10^{-7}$ to $10^{-8}$,  are less important than all other error sources (e.g., mal-trained perception network).

\section{Related Research}
Safety of AVs has attracted significant interest in academia and industry. Surprisingly, despite the serious effort on safety assurance, we did not identify studies close to the work we present in this paper.
Despite the fact that state-based models (Continuous Time Markov Chains, Stochastic Petri Nets, etc.) have been used in safety assessment of cyber-physical systems extensively, including to address safety in combination with other non-functional system properties (e.g. security and performance, including work by some of the present authors \cite{Bloomfield2017,Popov2017}). Earlier in this paper, we have already indicated a number of relevant references. Below we briefly summarize work of which we are aware using different methods for quantitative safety assessment. 

Hughes et al.~\cite{Hughes2015} provide an extensive survey of the models used for road safety assessment; they classify these using 7 categories. The model presented in this paper would belong to two of the listed categories – “system” model and “mathematical” model. 
In a number of publications, e.g. ~\cite{Althoff2009,Althoff2009b,Althoff2011} Althoff et al. use stochastic state-based models. The focus of these models is on estimating ("predicting”) the probability of accident in a specific situation (e.g., on a specific road scenery with a set of other vehicles and pedestrians). These works capture the effect of uncertainty about the trajectories that an ego vehicle would take and also study how the accuracy of estimates is affected by different probabilistic formalisms (e.g. a discrete time Markov chains and Monte Carlo simulation). These models can be seen as complementary to our work: they offer a credible way to undertake parameter estimation of the probabilities of accident, which we use in our work, but assume as given (or apply sensitivity analysis to see their impact on the chosen measure of system safety). A large number of similar works on “short-term” predictions of accidents are found in the literature; they vary in the assumptions made about the uncertainty captured and in the efficiency of the solutions.
To decide whether a vehicle is in fact proving as safe as argued on the basis of predictive models, one needs to look at statistic of the amount of driving and the number of accidents. However, as mentioned in the introduction, to demonstrate bounds on accident rates that would be tolerable to society would require observing amounts of accident-free operation that are commonly claimed to be unaffordable. This caused  significant interest both in academia and in industry for the paper by Kalra et al~\cite{Kalra2016}, often referred to as “driving to safety”. Their observation repeat those published by several authors (e.g.~\cite{Butler1991,Littlewood1993}) in the 1990s about similarly “ultra-high” dependability requirements in civil aviation, given the comparatively small amounts of safe test operation that were considered affordable before commercial operation: if one wants substantial confidence in the required safety level having been accomplished, one needs to derive it, in part at least, from other sources than observed safe operation.  Kalra et al. used as their required bound the average rate of fatal accidents per mile measured for human drivers, from data provided by NHTSA (National Highway Traffic Safety Administration) in the USA. Their paper triggered  responses to clarify how such prior evidence could be rationally combined with observed safe operation by Bayesian reasoning (in this case, ``Conservative Bayesian Inference'', or CBI, a way of simplifying Bayesian reasoning without risking errors that would overestimate safety)~\cite{Salako2021,Salako2023}.

It is common practice to start operation of a new AV type on a small scale only, and ramp it up progressively if no surprises occur that call the claim of ``ultra-high dependability'' for the type into question.
Bishop et al~\cite{Bishop2021} argue that this ``confidence bootstrapping'' practice is inevitable, because such claims have a non-negligible probability of being wrong, and they offer a CBI-based method for limiting risk during this incremental deployment. %

Among the effort on safety assessment by AV vendors we cite for example Waymo: Favaro et al. spell out ~\cite{Favaro2023} Waymo’s approach to safety assessment: safety, or 
“absence of unreasonable risk”, 
is assessed in terms of safety analyses at 3 levels – architecture, behavioral and operational. According to this layered approach, our work covers all three layers – we account for imperfection of components (such as driving policy and perception incl. safety monitors as explained in Section~II), we account for hazards due to AV being exposed to road hazards and finally, in operation some of the hazards may or may not escalate to accidents. In other publications (available via Waymo’s web-site at www.waymo.com/safety) the safety team provides further details including testing, and use the miles driven and accident records to defend the claim of absence of unreasonable risk. For instance,~\cite{Kusano2023} provides data on how the ``Waymo Driver'', a SAE Level 4 Automated Driving System, has compared to human drivers in terms of crashes and ``property damage''. They report that it had lower frequencies of various categories of minor accidents than human drivers, and that the superiority is statistically significant for some of the locations included in the 7.0+ millions miles in their statistics. While the amount of driving in the statistics is not enough for claiming improvements in the frequency of more serious, e.g. fatal, accidents, they claim that these statistics support Waymo's approach to safety. 
\section{Conclusion}
Assessing and demonstrating sufficient safety of an AV quantitatively is an open research challenge. At the same time, the trends in standardization and legal regulation highlight a goal that the probability of an AV causing an accident needs to be significantly lower compared to a human driver. Therefore, it is of utmost importance to be able to quantitatively estimate the probability of a catastrophic failure.

The challenge is that this failure probability is impacted by several factors. On the one hand, there are imperfections in the AV processing stack, e.g., errors in the perception or planning system, that can cause an accident. On the other hand, there are hazards imposed by the environment. 
In addition, the two factors influence each other, as e.g., an especially hazardous environment might be especially challenging for the AV perception system and misbehavior of the latter may, in turn, create hazardous conditions.  

In this work we proposed a formal model with significant complexity that allows one to estimate the final system failure probability (i.e., of accidents of different severity) based on the aforementioned influential factors. In other words, our model quantifies SOTIF performance limitations of the AV, related to situational awareness. The purpose here is to allow a designer to allocate reliability requirements to subsystems and functions and thus to drive system design; and to possibly produce an initial claim for the safety of the vehicle, to be combined with additional empirical evidence.
This requires an analyst to (a) detail the model, or multiple models for predictions under different conditions, from our description of a single hazard type and eight high-level states; (b) estimate parameter values.
For both needs, there will be tradeoffs between the degree of detail and of accuracy desired and the practical problems in dealing with complex modeling and detailed data analysis.
In particular, we acknowledge that it is still an open question how the required parameters can be estimated in a reliable manner. Nevertheless, we showed that initial estimates are possible for some of the parameters from available data. We are confident that some other parameters can be accurately estimated via tailored data acquisition, e.g., using fleet information obtained by OEMs from the sensors in their vehicles. In particular, these recordings will allow updating the data we used according to how driving habits change between geographical regions and over time (in part due to the very introduction of AVs). At that point, the model[s] will help to show whether any parameters remain hard to estimate empirically because they relate to very rare, or hard to observe, events. 

Besides the actual estimation of the accident rate, we also showed the usefulness of the model for sensitivity analysis. This allows one to quantify the impact of certain failures on the overall system (among them the failure rates of perception system and driving policy), to identify critical parameters and adapt data collection, analysis, and potentially the AV design, accordingly. 

Our evaluation 
showed that some of the parameters used in our analysis would have, in our example,  only minor effects, while others must be considered carefully to achieve safe driving. Perception and planning errors have, as expected, an important influence on the accident rate. Information like this can be used by AV manufacturers that employ such models early in their design process, with their own detailed models and data, to adapt their architectures accordingly to achieve the desired level of safety.


\appendices
\begin{table}[t!]
\centering
\begin{tabular}{|l|c|l|c|}
\hline
State & Count \# & Model variable & Rate [h$^{-1}$] \\
\hline
IS\_S\_B2IS\_S\_NB & 276 & rate\_IS\_S\_B2IS\_S\_NB & 646.87 \\
IS\_S\_B2IS\_F\_B & 0 & rate\_IS\_S\_B2IS\_F\_B & \\
IS\_S\_B2IS\_F\_NB & 91 & rate\_IS\_S\_B2IS\_F\_NB & 743.98 \\
IS\_S\_B2F\_S\_B & 45 & rate\_IS\_S\_B2F\_S\_B & 848.17 \\
IS\_S\_B2F\_S\_NB & 73 & rate\_IS\_S\_B2F\_S\_NB & 511.62 \\
IS\_S\_B2F\_F\_B & 0 & rate\_IS\_S\_B2F\_F\_B & \\
IS\_S\_B2F\_F\_NB & 101 & rate\_IS\_S\_B2F\_F\_NB & 573.80 \\
IS\_S\_NB2IS\_S\_B & 84 & rate\_IS\_S\_NB2IS\_S\_B & 601.59 \\
IS\_S\_NB2IS\_F\_B & 0 & rate\_IS\_S\_NB2IS\_F\_B & \\
IS\_S\_NB2IS\_F\_NB & 337 & rate\_IS\_S\_NB2IS\_F\_NB & 678.65 \\
IS\_S\_NB2F\_S\_B & 24 & rate\_IS\_S\_NB2F\_S\_B & 328.93 \\
IS\_S\_NB2F\_S\_NB & 423 & rate\_IS\_S\_NB2F\_S\_NB & 324.69 \\
IS\_S\_NB2F\_F\_B & 0 & rate\_IS\_S\_NB2F\_F\_B & \\
IS\_S\_NB2F\_F\_NB & 385 & rate\_IS\_S\_NB2F\_F\_NB & 207.38 \\
IS\_F\_B2IS\_S\_B & 33 & rate\_IS\_F\_B2IS\_S\_B & 1164.7 \\
IS\_F\_B2IS\_S\_NB & 0 & rate\_IS\_F\_B2IS\_S\_NB & \\
IS\_F\_B2IS\_F\_NB & 14 & rate\_IS\_F\_B2IS\_F\_NB & 1680 \\
IS\_F\_B2F\_S\_B & 22 & rate\_IS\_F\_B2F\_S\_B & 728.83 \\
IS\_F\_B2F\_S\_NB & 11 & rate\_IS\_F\_B2F\_S\_NB & 430.43 \\
IS\_F\_B2F\_F\_B & 76 & rate\_IS\_F\_B2F\_F\_B & 986.54 \\
IS\_F\_B2F\_F\_NB & 110 & rate\_IS\_F\_B2F\_F\_NB & 822.71 \\
IS\_F\_NB2IS\_S\_B & 43 & rate\_IS\_F\_NB2IS\_S\_B & 936.29 \\
IS\_F\_NB2IS\_S\_NB & 52 & rate\_IS\_F\_NB2IS\_S\_NB & 1109.88 \\
IS\_F\_NB2IS\_F\_B & 2 & rate\_IS\_F\_NB2IS\_F\_B & 1800 \\
IS\_F\_NB2F\_S\_B & 137 & rate\_IS\_F\_NB2F\_S\_B & 560.88 \\
IS\_F\_NB2F\_S\_NB & 115 & rate\_IS\_F\_NB2F\_S\_NB & 918.64 \\
IS\_F\_NB2F\_F\_B & 163 & rate\_IS\_F\_NB2F\_F\_B & 963.55 \\
IS\_F\_NB2F\_F\_NB & 2599 & rate\_IS\_F\_NB2F\_F\_NB & 948.25 \\
F\_S\_B2IS\_S\_B & 139 & rate\_F\_S\_B2IS\_S\_B & 940.01 \\
F\_S\_B2IS\_S\_NB & 238 & rate\_F\_S\_B2IS\_S\_NB & 438.63 \\
F\_S\_B2IS\_F\_B & 0 & rate\_F\_S\_B2IS\_F\_B & \\
F\_S\_B2IS\_F\_NB & 107 & rate\_F\_S\_B2IS\_F\_NB & 439.22 \\
F\_S\_B2F\_S\_NB & 2234 & rate\_F\_S\_B2F\_S\_NB & 465.98 \\
F\_S\_B2F\_F\_B & 1 & rate\_F\_S\_B2F\_F\_B & 1080 \\
F\_S\_B2F\_F\_NB & 340 & rate\_F\_S\_B2F\_F\_NB & 604.64 \\
F\_S\_NB2IS\_S\_B & 42 & rate\_F\_S\_NB2IS\_S\_B & 219.02 \\
F\_S\_NB2IS\_S\_NB & 582 & rate\_F\_S\_NB2IS\_S\_NB & 228.71 \\
F\_S\_NB2IS\_F\_B & 0 & rate\_F\_S\_NB2IS\_F\_B & \\
F\_S\_NB2IS\_F\_NB & 325 & rate\_F\_S\_NB2IS\_F\_NB & 264.45 \\
F\_S\_NB2F\_S\_B & 604 & rate\_F\_S\_NB2F\_S\_B & 355.80 \\
F\_S\_NB2F\_F\_B & 1 & rate\_F\_S\_NB2F\_F\_B & 568.42 \\
F\_S\_NB2F\_F\_NB & 2066 & rate\_F\_S\_NB2F\_F\_NB & 223.71 \\
F\_F\_B2IS\_S\_B & 148 & rate\_F\_F\_B2IS\_S\_B & 549.09 \\
F\_F\_B2IS\_S\_NB & 4 & rate\_F\_F\_B2IS\_S\_NB & 310.79 \\
F\_F\_B2IS\_F\_B & 103 & rate\_F\_F\_B2IS\_F\_B & 616.291 \\
F\_F\_B2IS\_F\_NB & 67 & rate\_F\_F\_B2IS\_F\_NB & 551.10 \\
F\_F\_B2F\_S\_B & 1584 & rate\_F\_F\_B2F\_S\_B & 676.71 \\
F\_F\_B2F\_S\_NB & 41 & rate\_F\_F\_B2F\_S\_NB & 319.02 \\
F\_F\_B2F\_F\_NB & 1164 & rate\_F\_F\_B2F\_F\_NB & 609.87 \\
F\_F\_NB2IS\_S\_B & 97 & rate\_F\_F\_NB2IS\_S\_B & 361.87 \\
F\_F\_NB2IS\_S\_NB & 101 & rate\_F\_F\_NB2IS\_S\_NB & 273.31 \\
F\_F\_NB2IS\_F\_B & 161 & rate\_F\_F\_NB2IS\_F\_B & 209.77 \\
F\_F\_NB2IS\_F\_NB & 2170 & rate\_F\_F\_NB2IS\_F\_NB & 202.34 \\
F\_F\_NB2F\_S\_B & 643 & rate\_F\_F\_NB2F\_S\_B & 252.79 \\
F\_F\_NB2F\_S\_NB & 724 & rate\_F\_F\_NB2F\_S\_NB & 361.58 \\
F\_F\_NB2F\_F\_B & 2870 & rate\_F\_F\_NB2F\_F\_B & 170.61 \\
\hline
\end{tabular}
\caption{Summary of the data extracted from our evaluation dataset}
\label{table:state_transition}
\vspace{-0.4cm}
\end{table}

\section{}

An extended version of Table~\ref{table:state_transition_short} is provided in Table~\ref{table:state_transition}. One notices that 7 of the transitions listed in Table~\ref{table:state_transition} never occurred in the dataset. A further 4 transitions occurred very rarely, between 1 and 4 times in the dataset. This significant disparity in the likelihood of transitions to different states is initially ignored and the model is built using only the transition rates. These together with “competing risks” models are sufficient to construct a complete continuous-time Markov chain model.

\ifCLASSOPTIONcaptionsoff
  \newpage
\fi



\bibliographystyle{IEEEtran}
\bibliography{main.bib}

\begin{IEEEbiography}[{\includegraphics[width=1in,height=1.25in,clip,keepaspectratio]{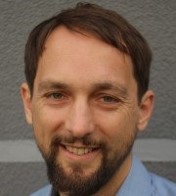}}]{Cornelius Buerkle}
is a Research Scientist at Intel Labs, where he focuses on research on dependable autonomous systems, with focus on safe perception. Cornelius has more than a decade of experience in the automotive industry and worked in research on autonomous systems and on the pre-development of Driver Assistance Systems. He graduated at the University of Karlsruhe (KIT) and holds a Diploma in Computer Science.
\end{IEEEbiography}
\vspace{-1cm}
\begin{IEEEbiography}[{\includegraphics[width=1in,height=1.25in,clip,keepaspectratio]{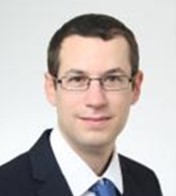}}]{Fabian Oboril}
is a researcher at Intel Labs with focus on safe, risk-aware approaches for autonomous robots, including automated vehicles, in particular enabling safe perception solutions. Prior to Intel, Fabian was a PostDoc at the Karlsruhe Institute of Technology, where he worked on spintronics and dependable computing. Since 2016, he holds a PhD in computer science for his research in dependability-aware microprocessor design.
\end{IEEEbiography}
\vspace{-1cm}
\begin{IEEEbiography}[{\includegraphics[width=1in,height=1.25in,clip,keepaspectratio]{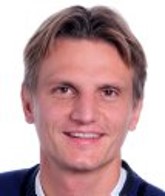}}]{Michael Paulitsch}
Head of Extended Reality Lab @ Intel Labs | Senior AI Researcher. He holds a PhD degree from the Technical University Vienna. He has a proven track record of successfully leading technology innovation and transferring business-critical technology to key products in the IT, transport and aerospace and defense industries. He is adept in agile management practices, multi-million cost center management, and acquisition, and leading research projects. 
He has a strong interest in R\&D, automation and optimization using technology, with a focus on novel AI/ML models. He loves working with excellent teams and is driven by change via transfer of leading-edge research to products and customers, creating new business value through software ecosystems, optimizing processes \& systems.
\end{IEEEbiography}
\vspace{-1cm}
\begin{IEEEbiography}[{\includegraphics[width=1in,height=1.25in,clip,keepaspectratio]{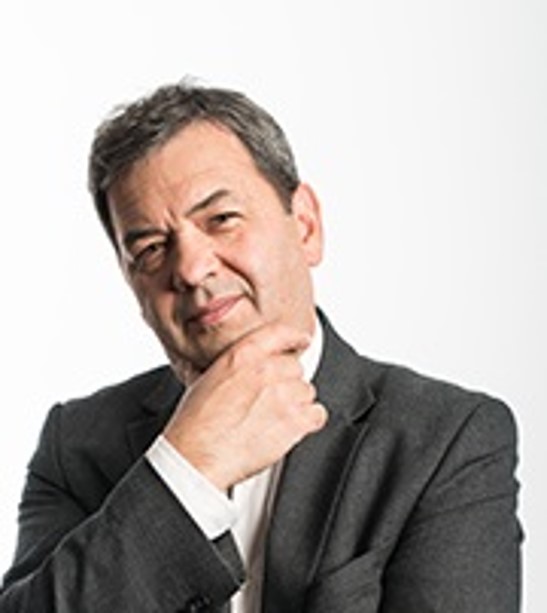}}]{Peter Popov}
(a member of IEEE). Peter is Reader in Systems Dependability at City St George's, University of London. 
His research interests include dependability assessment of computer-based systems. He is known for his work on probabilistic modeling for dependability assessment, especially of fault tolerant software using “design diversity”, modeling large interdependent critical infrastructures and the impact of cyber-attacks on safety critical systems. 
His current research is focused on probabilistic modeling for safety assessment of autonomous vehicles and on cyber-resilience of Connected and Autonomous Vehicles (CAVs). 
He is a member of the IFIP WG 10.4 on Dependable Computing and Fault Tolerance and of the WG on Safety and Security set up in by the UK National Cyber-Security Centre (NCSC) to develop “Safety and Security: Code of Practice”.
\end{IEEEbiography}
\vspace{-1cm}
\begin{IEEEbiography}[{\includegraphics[width=1in,height=1.25in,clip,keepaspectratio]{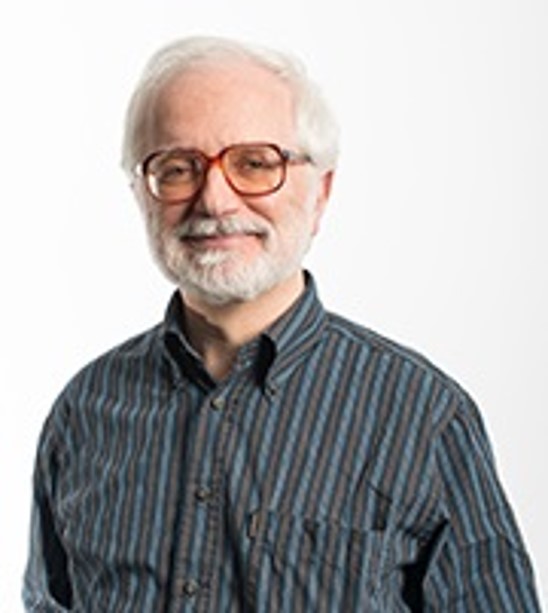}}]{Lorenzo Strigini}
(Member, IEEE), is Professor of Systems Engineering in the  Centre for Software Reliability (CSR) at City St George's, University of London. His research focuses on probabilistic reasoning on dependability, to help design as well as to assess working systems; with a focus on diversity and defense-in-depth. His publications cover both theory results and applied studies, in areas spanning nuclear safety systems, acceptance decisions for safety-critical systems, and interacting safety and security concerns. He has been a PI for various national and international projects in these areas, and a consultant to industry; he is a member of IFIP WG 10.4 and until recently the Associate Editor in Chief for IEEE Transactions on Dependable and Secure Computing
\end{IEEEbiography}




\end{document}